\documentclass[lettersize,journal]{IEEEtran}
\usepackage{amsmath,amsfonts}
\usepackage{algorithm}
\usepackage{algpseudocode}
\usepackage{array}
\usepackage[caption=false,font=normalsize,labelfont=sf,textfont=sf]{subfig}
\usepackage{textcomp}
\usepackage{stfloats}
\usepackage{url}
\usepackage{verbatim}
\usepackage{graphicx}
\usepackage{cite}

\usepackage{tablefootnote}

\usepackage{hyperref}  
\usepackage{amssymb}
\usepackage{amsmath,amsthm}
\usepackage{tikz}
\usepackage{mathtools}
\usepackage{thmtools}
\usepackage{graphicx}
\usepackage{booktabs}       
\usepackage{multirow}
\usepackage{MnSymbol}
\usepackage{caption}
\graphicspath{ {plots/} } 
\newcommand{\indep}{\perp \!\!\! \perp}

\theoremstyle{definition}
\newtheorem{definition}{Definition}
\newtheorem{example}{Example}

\begin{document}

\title{Entry Dependent Expert Selection in Distributed Gaussian Processes Using Multilabel Classification}

\author{Hamed Jalali, Gjergji Kasneci
\thanks{Authors are with the Data Science and Analytics Research (DSAR) group at the University of Tübingen, 72076 Tübingen, Germany (E-mail: hamed.jalali@uni-tuebingen.de; gjergji.kasneci@uni-tuebingen.de).\\
}
}



\maketitle

\begin{abstract}
By distributing the training process, local approximation reduces the cost of the standard Gaussian Process. An ensemble technique combines local predictions from Gaussian experts trained on different partitions of the data. Ensemble methods aggregate models' predictions by assuming a perfect diversity of local predictors. Although it keeps the aggregation tractable, this assumption is often violated in practice. Even though ensemble methods provide consistent results by assuming dependencies between experts, they have a high computational cost, which is cubic in the number of experts involved. By implementing an expert selection strategy, the final aggregation step uses fewer experts and is more efficient. However, a selection approach that assigns a fixed set of experts to each new data point cannot encode the specific properties of each unique data point. This paper proposes a flexible expert selection approach based on the characteristics of entry data points. To this end, we investigate the selection task as a multi-label classification problem where the experts define labels, and each entry point is assigned to some experts. The proposed solution's prediction quality, efficiency, and asymptotic properties are discussed in detail. We demonstrate the efficacy of our method through extensive numerical experiments using synthetic and real-world data sets.
\end{abstract}


\section{Introduction}
\IEEEPARstart{G}{aussian}  processes (GPs) \cite{Rasmussen} are interpretable and powerful Bayesian non-parametric methods for non-linear regression. A Gaussian process is a stochastic process where every finite collection of those random variables has a multivariate Gaussian distribution.  By applying Bayes' theorem for inference, the posterior predictive distribution of a GP is the best linear unbiased estimator (BLUE) under the assumed model and provides proper quantification of the prediction error uncertainty. GPs do not need restrictive assumptions of the model and can estimate complex linear and non-linear structures. While GPs are extensively used in practical cases \cite{Deringer, Zeng2020, Denzel, Gramacy, Le, Tobar}, their cubic training and quadratic prediction costs~\footnote{I.e., in the size of the training set.} limit their application to big data use cases~\cite{Deisenroth}.

For GP regression, the major computational hurdle is the need to estimate the kernel inversion and determinant, which is prohibitively expensive when $n$ is large. Because of this issue, GPs are typically restricted to relatively small training data sets in the range of $\mathcal{O}(10^4)$.

To reduce computational expense, sparse approximation techniques use a subset of the training sets (called inducing points) and \textit{Nyström} approximation to estimate posterior distributions \cite{Titsias, Hensman, Cheng, Burt}. In this case, this approach provides a full probabilistic model and appropriate predictions based on the Bayesian framework. Despite its advantages, this method cannot handle large data sets since its capacity is limited by the number of inducing points~\cite{Bui, Moore}.

Unlike the sparse approximation, which uses only the inducing points, the prominent distributed Gaussian processes (DGPs) use the full training set. This method uses \emph{centralized distributed learning}, which means that the training data is partitioned into numerous subsets, the local inference is conducted for each partition separately, and then the local estimations are combined through ensemble learning~\cite{Barral, Verbraeken_2020, Liu_Huang}. A local GP with expertise in a particular partition is called an expert. Experts share the same hyper-parameters, thus accounting for implicit regularisation and encountering overfitting~\cite{Deisenroth, Liu2020}.

In a DGP, the \emph{conditional independence} (CI) assumption between partitions (i.e., between experts given the target) allows for factorizing the global posterior distribution as a product of local distributions. While this assumption reduces the computational cost, it results in inconsistencies and suboptimal solutions~\cite{Szabo} caused by the partitioning of the data set, such that 
when $N\to \infty$, the CI-based posterior approximations do not converge to the full GP posterior.

Relaxing the independence assumption raises the aggregation's theoretical properties. If the experts' predictions are assumed to be random variables, their relative correlations define dependencies between experts. The aggregated posterior distribution, in this case, provides high-quality forecasts and is capable of returning consistent results \cite{Rulliere, Bachoc, jalali2022}. However, solutions that deal with the consistency problem suffer from extra computational costs induced by the need to find the inverse of the covariance matrix between experts for each test point. It means the complexity of this model cubically depends on the number of experts (say $M$), and therefore it can become computationally prohibitive when $M$ is large. 

Few works have considered boosting the efficiency of dependency-based aggregation. In~\cite{jalali2021,jalali2021_arxiv}, authors discuss complexity reduction as an expert selection scenario that excludes a subset of original experts and considers only the valuable experts in the aggregation. For this purpose, the precision matrix of the experts' predictions is estimated using a Gaussian graphical model (GGM). Experts are the nodes in the obtained undirected sparse graph, and their interactions are the edges. The nodes with fewer interactions are defined as unimportant experts and excluded from the model. This approach can lower the complexity and provide a good approximation for the original estimator. However, it is not flexible concerning new entries, and the selected experts are fixed for all test points. If new entry data points have specific behavior or are close to excluded partitions, the prediction error increases. 

The critical contribution of our work lies in selecting a subset of local experts for each new data point using multi-label classification. Unlike the static expert selection by GGM~\cite{jalali2021}, the proposed method does not assign a fixed set of local experts for all test points. A dynamic and flexible mechanism for each new observation designates related experts to provide local predictions. Multi-label classification~\cite{Herrera2016} is a generalization of multi-class classification, where multiple labels may be assigned to each instance. It originates from the investigation of the text categorization problem, where each document may belong to several predefined topics simultaneously. 

To transform the distributed learning case into a multi-label classification, the indices of the partitions/experts are the labels/classes. The task is to assign some experts to a new data point. Multi-class classification problem would select an appropriate expert for predicting and would lead to a local approximation with only one expert per test point. This one-expert inductive model, however, produces discontinuous separation boundaries between sub-regions and therefore is not a proper solution for quantifying uncertainties~\cite{Park, Liu2020}. 

Two algorithms can be adapted to assign experts to data points: k-nearest neighbors (KNN) and deep neural networks (DNN). For the first one, we use the centroid of the partition as a substitute for the corresponding local expert. By estimating the distance between a new entry point and the centroids, we can find its $K$ nearest neighboring experts. Due to the properties of the Gaussian process experts, if a test point is close to a GP expert, the expert can provide a reliable prediction for that test point.

For the second approach, we train a neural network with a soft-max output layer and log-loss (i.e., cross-entropy loss) using the train points and their related partition index that shows the partition they belong to. After training the DNN, we send a new test point through the network, and the experts with higher probability are assigned to this test point. Relative to consistent aggregation methods that use dependency information, our approach keeps all asymptotic properties of the original baseline and substantially provides competitive prediction performance while leading to better computational costs than other SOTA approaches, which use the dependency assumption. By extending the proposed method for CI-based ensembles, we can use it in federated learning problems, which do not consider dependencies between agents, see \cite{Blanchard_2017,data_2021}.

The structure of the paper is as follows. Section \ref{problem_setup} introduces the problem formulation and related works. The proposed model and inference process are presented in Section \ref{contribution}. Section \ref{discussion} discusses some associated details. Section \ref{experiments} shows the experimental results, and we conclude in Section \ref{conclusion}.

\section{Problem Set-up} \label{problem_setup}

\subsection{Gaussian Process}
We start with the basic non-linear regression problem $y=f(x)+\epsilon$, and the objective is to learn the latent function \textit{f} from a training set $\mathcal{P}=\{X,y\}$. Assume the training set contains $N$ observation, $X$ is a d-dimensional variable, $x\in R^d$, and $\epsilon$ is a zero-mean Gaussian noise $\epsilon \sim \mathcal{N}(0,\sigma^2)$.
The Gaussian process describes a prior distribution over the latent functions as $f \sim \mathcal{N}\left(0,k(x,x') \right)$, where $k(x,x')$ is the covariate function (kernel) with hyperparameters $\psi$, and $x,x' \in X$. The prior kernel is the  well-known squared exponential (SE) covariance function equipped with automatic relevance determination (ARD), 
\[k(x,x^{'})=s^2 \; \exp\left( -\frac{1}{2} \sum_{i=1}^d \frac{(x_i-x_i')^2}{\mathcal{T}_i} \right),\]
where $\sigma_f^2$ is the signal variance, and $\mathcal{T}_i$ is a correlation length scale parameter along the $i$-th dimension. Let $\tau=\{s^2,\mathcal{T}_1,\ldots,\mathcal{T}_d\}$, training the GP involves determining the hyperparameters $\theta=\{\sigma^2, \tau\}$ such that they maximise the related log-marginal likelihood,
\begin{equation} \label{eq:log_likelihood}
\log p(y|X)=-\frac{1}{2}y^T\mathcal{Z}^{-1}y - \frac{1}{2} \log|\mathcal{Z}|- \frac{N}{2} \log2\pi,
\end{equation}
where $\mathcal{Z}=k(X,X)+\sigma^2I$.
Optimization task in \eqref{eq:log_likelihood} scales as $\mathcal{O}(N^3)$ because it requires to calculate the inversion of the $N \times N$ matrix $\mathcal{Z}$. It inflicts limitations on the scalability of GPs, and the training step is time-consuming for large data sets.

\subsection{Local Approximation Gaussian Process}
The local approximation Gaussian process is a \textit{divide-and-conquer} approach, which partitions the training data set into $M$ subsets $\mathcal{P}^{'}= \{\mathcal{P}_1,\ldots,\mathcal{P}_M\}$ and trains standard GPs on these subsets. It is also called the distributed Gaussian process (DGP for short), which builds on distributing the training of the standard GP among several computing units. Let $X_i$ and $y_i$ be the input and output of subset $\mathcal{P}_i$. The related local GPs at each subset are called experts that are trained jointly and share a single set of hyperparameters $\theta=\{\sigma^2,\psi\}$ \cite{Deisenroth}. 

The local predictive distribution of the $i$-th expert $\mathcal{E}_i$ on a test set $X^*$ of size $N_t$ is $p_i(y^*|\mathcal{P}_i,X^*)\sim \mathcal{N}(\mu_i^*,\Sigma_i^*)$ with mean and covariance as:
\begin{align}
\mu_i^* &= k_{i*}^T(K_i+\sigma^2I)^{-1}y_i,\label{eq:local_mean} \\
\Sigma_i^* &= k_{**} - k_{i*}^T(K_i+\sigma^2I)^{-1}k_{i*}, \label{eq:local_var}
\end{align}
where $K_i=k(X_i,X_i)$, $k_{i*}=k(X_i,X^*)$, and $k_{**}=k(X^*,X^*)$. 

To divide the training data set $\mathcal{P}$ into $M$ partitions, two different strategies are used: \textit{random} and \textit{disjoint} partitioning. Although random partitioning is faster than disjoint partitioning, it has been widely accepted that disjoint partitioning can capture the local features of the data more accurately, see \cite{Liu2018, jalali2021}. Therefore, we assign training data to experts using a K-mean clustering approach in this work.

A DGP typically aggregates the local GP experts assuming perfect diversity between them which means they are conditionally independent, i.e., $\mathcal{E}_i \upmodels \mathcal{E}_j |y^*$ where $i,j \in \{1,\ldots,M\}$. Using CI assumption between experts $\{\mathcal{E}\}_{i=1}^M$ allows us to factorize the predictive distribution of a DGP over all local predictive distributions. That is to say, for a test input $x^*$  
\begin{equation} \label{eq:CI_ensemble}
p(y^*|\mathcal{P},x^*) \propto \prod_{i=1}^M p_i^{\beta_i}(y^*|\mathcal{P}_i,x^*).
\end{equation}
Equation \eqref{eq:CI_ensemble} shows that the aggregated predictive distribution can be defined as the product of local densities. The $\beta = \{\beta_1,\ldots,\beta_M\}$ describe the weights and influence of the experts. The most prevalent CI-based aggregation methods are product of experts (PoE) \cite{Hinton}, generalized product of experts (GPoE) \cite{Cao}, Bayesian committee machine (BCM) \cite{Tresp}, robust Bayesian committee machine (RBCM) \cite{Deisenroth} and generalized robust Bayesian committee machine (GRBCM) \cite{Liu2018}. 

\begin{figure}[!t] 
\centering
\includegraphics[width=\columnwidth]{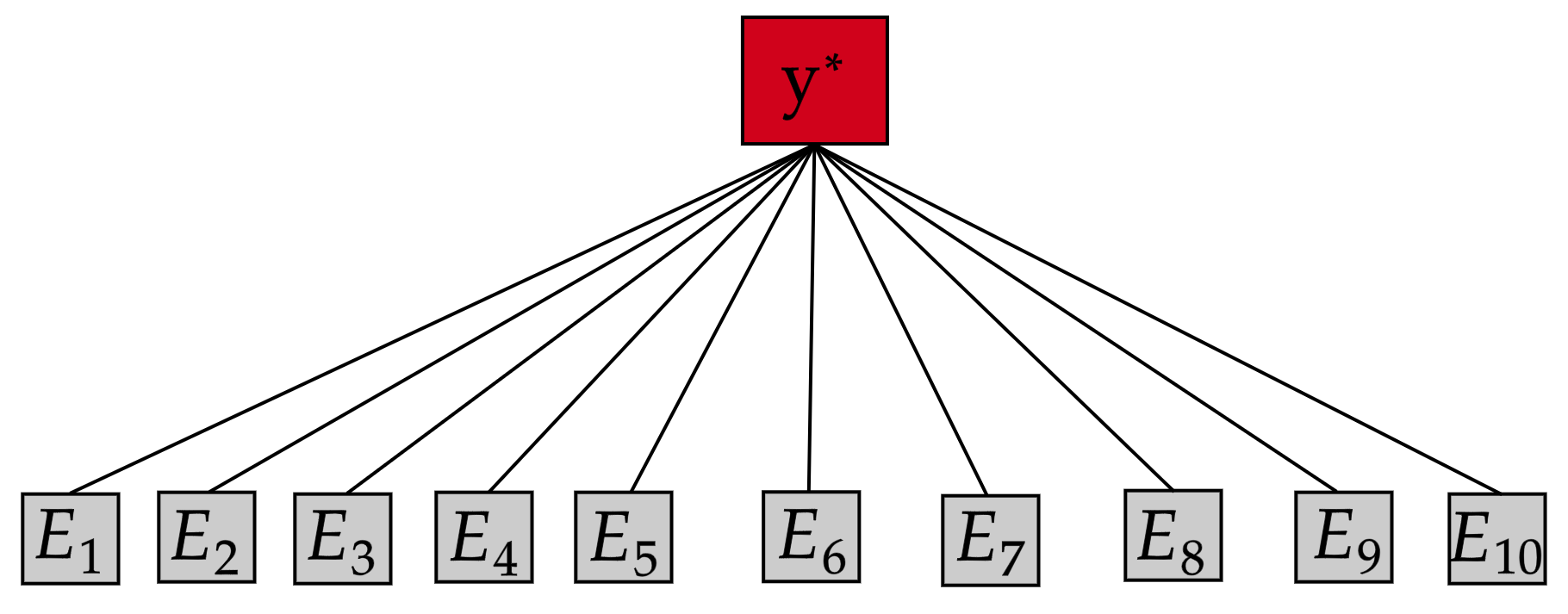}
\caption{\textbf{Computational graphs} of an aggregation based on conditional independence assumption between experts.}
\label{aggregation_graphs}
\vskip -0.1in
\end{figure}

Figure \ref{aggregation_graphs} depicts the computational graph of the DGP strategy. It reveals the aggregation based on conditional independence assumption between experts $\{\mathcal{E}_1,\ldots,\mathcal{E}_{10} \}$. The CI assumption means two local experts $\mathcal{E}_i$ and $\mathcal{E}_j$ are connected only via the target variable $y^*$, i.e. $\mathcal{E}_i \indep \mathcal{E}_j \mid y^*$. Thus, there is no interaction between experts, and they can not affect each other.

\subsection{Beyond Conditional Independence Assumption}
Ensemble methods extensively employ the conditional independence assumption for regression and classification problems~\cite{Moreira, Parisi}. Although this assumption reduces the prediction cost of DGPs, it generally leads to a sub-optimal solution and their related predictions are not accurate enough~\cite{Jaffe}. In classification, modeling dependencies between classifiers has been considered in several works, for example, in~\cite{Donmez,  Platanios, Jaffe}. However, few works have considered modeling expert dependencies in local approximation GPs. The nested pointwise aggregation of experts (NPAE) method~\cite{Rulliere, Bachoc} defines an estimator using the interactions between experts and the target variable $y^*$. 

For a given test point $x^*\in X^*$, assume the vector $\mu^*(x^*)=[\mu_1^*(x^*),\ldots,\mu_M^*(x^*)]^T$ contains the centered predictions of $M$ local GP experts $\mathcal{E}=\{\mathcal{E}_1,\ldots,\mathcal{E}_M\}$, where $\mu_i^*(x^*), i=1,\ldots,M$ has been defined in \eqref{eq:local_mean}. Each of the local Gaussian experts $\mathcal{E}_i$ is a linear estimator because the related prediction $\mu^*_i$ is linear with respect to the observed values of the random variable $y_i$, i.e., $\mu_i^*=Q_i y_i$, where $Q_i =k^T_{i*}(K_i + \sigma^2 I)^{-1}$. Authors in \cite{Rulliere} assumed that $y_i$ in \eqref{eq:local_mean} has not yet been observed. It allows us to consider $\mu_i^*(x^*)$ as a \emph{random variable}. Therefore, the experts' dependencies can be investigated in two ways; the correlations between the experts' predictions and target variable, \ $Cov(\mu^*_i, y^*)$, and internal correlations between experts' predictions, \ $Cov(\mu^*_i,\mu^*_j)$ where $i,j=1,\ldots,M$. The analytical explanation of both covariances can be defined as:

\begin{equation} \label{eq:cross_cov}
    Cov(\mu_i^*, y^*)=cov(Q_i y_i,y^*)=Q_i k(X_i,X^*)
\end{equation}
\begin{equation} \label{eq:auto_cov}
    Cov(\mu_i^*, \mu_j^*)=cov(Q_i y_i,Q_j y_j)=Q_i k(X_i,X_j) Q_j^T
\end{equation}

For a test point $x^*\in X^*$, the point-wise covariances are defined as $r(x^*)=Cov\left(\mu^*(x^*), y^*(x^*)\right)$ and $R(x^*)=Cov\left(\mu^*(x^*),\mu^*(x^*)\right)$, where $r(x^*)$ is a $M\times1$ vector and $R(x^*)$ is a $M\times M$ matrix. The joint distribution of random variables $(y^*,\mu_1^*,\ldots,\mu_M^*)$ is a multivariate normal distribution. This arises from the fact that every vector of linear combinations of normally distributed observations is itself a Gaussian vector. This fact is used to define the predictor $y_A^*(x^*)$ of $y^*(x^*)$ which aggregates variables $\mu_i^*(x^*), i=1,\ldots,M,$ and leads to the subsequent aggregation:

\begin{definition}[\textbf{Dependency-Based Aggregation}]
The aggregated predictor for the test point $x^*$ and local predictions $\mu_1^*(x^*),\ldots,\mu_M^*(x^*)$ is defined as
\begin{equation} \label{eq:npae}
y_A^*(x^*)=r(x^*)^T R(x^*)^{-1}\mu^*(x^*).
\end{equation}
\end{definition}
The NPAE estimator in \eqref{eq:npae} provides high-quality predictions. It is straightforward to show that this linear estimator is the \textit{best linear unbiased predictor} (BLUP), see \cite{Rulliere}.

\begin{example} [\textbf{Concrete  Data Set}] \label{example_1}
\textit{Concrete Compressive Strength} \footnote{\url{https://archive.ics.uci.edu/ml/datasets/concrete+compressive+strength}} data set contains 1030 observations of 9 attributes (8 independent variables and one response variable). We use $90\%$ of the observations for training and the rest for testing, where disjoint partitioning is used to divide the data set into 5, 7, and 10 subsets. The prediction quality of CI and dependency-based aggregations are compared with standard GP. The quality of predictions is evaluated in two ways, standardized mean squared error (SMSE) and the mean standardized log loss (MSLL). The SMSE measures the accuracy of the prediction mean, while the MSLL evaluates the quality of predictive distribution~\cite{Rasmussen}.
\end{example}

\begin{figure}[t!]
     \centering
     \subfloat[SMSE \label{example_smse}]{
         \centering
         \includegraphics[width=0.95\linewidth]{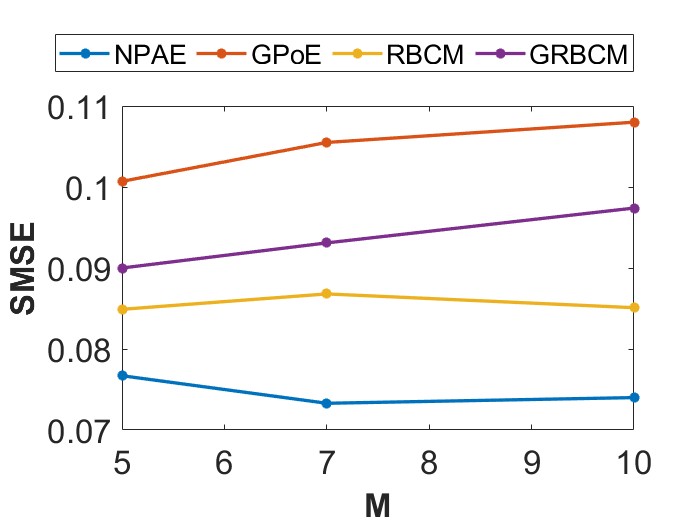}
     }
     
     \subfloat[MSLL \label{example_msll}]{
         \centering
         \includegraphics[width=0.95\linewidth]{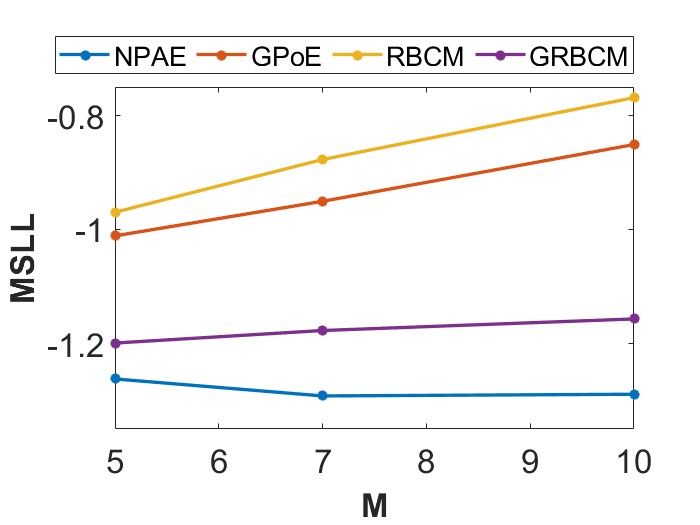}
     }
    \caption{\textbf{Ablation experiment1.} Prediction quality of different aggregation baselines for the \textit{Concrete} data set.}
    \label{fig:example_1}
\end{figure}

Figure \ref{fig:example_1} shows the prediction quality of available baselines for the \textit{Concrete} data set when the number of experts (M) changes. NPAE, which uses the dependencies between experts, provides consistently better results, confirming that modeling the experts' interactions improves the quality of the aggregated predictive distribution. Especially, the quality of NPAE is not sensitive to changes in the number of experts (M). On the other hand, increasing M (reducing the size of the sub-partitions) lowers the prediction accuracy of existing CI-based DGP methods.

\subsection{Asymptotic Properties}
 
Conventional DGP baselines suffer from \textit{inconsistency}. Since the local experts are trained on separated partitions, the aggregation produces inconsistent predictions that can not converge to the standard GP. Several works have investigated the asymptotic properties of the CI-based ensembles and confirmed the inconsistent and overconfident predictions of the PoE and (R)BCM methods. Besides, the GPoE with normalized equal weights~\cite{Deisenroth} conservatively converges to the full GP distribution when $N \to \infty$~\cite{Szabo, Bachoc}. However, the authors in ~\cite{Liu2018} showed that the GPoE produces consistent predictions using random partitioning under some mild assumptions. 

The generalized robust Bayesian committee machine (GRBCM)~\cite{Liu2018} introduces a base (global) expert and considers the covariance between the base and other local experts, which, under some mild assumptions, can provide consistent results using both random and disjoint partitioning. However, it still uses the CI-based aggregation in the RBCM method and sometimes yields poor results, particularly when the data is randomly partitioned.

The point-wise NPAE method is capable of providing consistent results. It benefits from both dependencies forms in Equations \eqref{eq:cross_cov} and \eqref{eq:auto_cov}, and the aggregated predictor in \eqref{eq:npae} produces high-quality predictions employing the properties of conditional Gaussian distribution. Estimating the inverse of the internal correlation $R(x^*)$ leads to two issues: the existence of the inversion matrix and computational cost. Using matrice's pseudo-inverse can solve the first issue, but the second complicates employing the NPAE for large data sets. Calculating the inverse of the $M\times M$ matrix $R(x^*)$ has cubic time complexity in the number of local experts at each test point $x^* \in X^*$. Therefore the aggregation cost is $\mathcal{O}(N_t M^3)$, which is not an efficient solution for complex real-world data sets with large $M$ and $N_t$ values.

In the next section, we propose a new expert selection approach using the multi-label classification model to assign test points to some proper experts instead of using the full experts set that modifies the aggregation estimator in \eqref{eq:npae}.

\section{Expert Selection in Local Approximation GPs} \label{contribution}

The current DGP baselines rely on experts' weighting to quantify the experts' importance. In~\cite{Deisenroth}, the authors discussed different weights for local experts and came to the conclusion that they can not outperform the linear mixture weights based on experts' correlations defined in \eqref{eq:npae}. However, despite the high accuracy of the NPAE aggregation, its computational cost is a challenge for using the method on large data sets.  

Expert selections can improve the performance of dependency-based aggregation in two ways. First, unrelated experts for a given data point can be excluded and only informative partitions can be considered to make the prediction. Second, mitigating the number of experts reduces the prediction cost and enables the ensemble to be used in large data sets.

\subsection{Expert Selection Using Graphical Models} \label{GGM_selection}

Gaussian graphical model (GGM) is the continuous form of pairwise Markov random fields. It assumes the nodes of an undirected graph are random variables, and the joint distribution of the random variables is multivariate Gaussian distribution with zero mean and precision matrix $\Omega$, $\mathcal{N}(0,\Omega^{-1})$. The elements of the precision matrix are the unknown parameters and show interactions between experts (edges in the graph).

Let $S$ be the sample covariance of local predictions $\mu^*$, i.e. $S=Cov(\mu^*)$. Then, the log-likelihood of the Gaussian multivariate distribution and precision matrix $\Omega$ is equal to $\log |\Omega| - tr(S \Omega)$. To estimate the precision matrix,  Graphical Lasso (GLasso) \cite{Friedman2008, Friedman2011} is an efficient inference algorithm that maximizes this likelihood subject to an element-wise $l_1$-norm penalty on $\Omega$. More precisely, the objective function is,

\begin{equation} \label{eq:glasso}
\widehat{\Omega}= \arg\max_{\Omega} \log |\Omega| - tr(S \Omega) - \lambda \left\Vert \Omega \right\Vert_1,
\end{equation}
where the estimated expert network is then given by the non-zero elements of $\widehat{\Omega}$. 

Modeling the dependency in distributed learning by GGMs has been studied in~\cite{jalali2020, jalali2022}, where the precision matrix encodes the interactions between experts. The authors in~\cite{jalali2020} used the GLasso algorithm to detect dependencies between Gaussian experts and identify clusters of strongly dependent experts. Besides, expert selection by GGM has been proposed and investigated in~\cite{jalali2021_arxiv,jalali2021}, which divides the experts into important and unimportant experts and excludes the unimportant experts in the final aggregation. The strength of an expert's interactions in the related undirected graph defines the expert's importance. 

\begin{definition}[\textbf{GGM-related Expert Importance}] \label{def.importance}
The importance of expert $\mathcal{E}_i$ based on the estimated precision matrix $\widehat{\Omega}$ is defined as $\mathcal{I}_i=\sum_{j=1, j\neq i}^M |\widehat{\Omega}_{ij}|$. 
\end{definition}
According to the interactions, the GGM-related expert selection task uses the first K experts in the descending sorted importance set $\mathcal{I}=\left\{\mathcal{I}_{i_1},\mathcal{I}_{i_2}, \ldots, \mathcal{I}_{i_M} \right\}$, leading to a new expert set. $\mathcal{M_{GGM}}=\{\mathcal{M^G}_1,\ldots,\mathcal{M^G}_K\}$ and $K< M$. The number of selected experts is defined as $K= \alpha \times M$, where $\alpha$ is a hyperparameter that indicates the percentage of original experts selected for the final aggregation. The experts in $\mathcal{M_{GGM}}$ are fixed and used for prediction at any new entry point. The GGM-based aggregation can provide consistent results, and its predictive distribution is a consistent approximation of the unbiased estimator in \eqref{eq:npae} \cite{jalali2021}.

\begin{figure}[t!]
     \centering
     \subfloat[GGM \label{fig:ggm}]{
         \includegraphics[width=0.9\columnwidth]{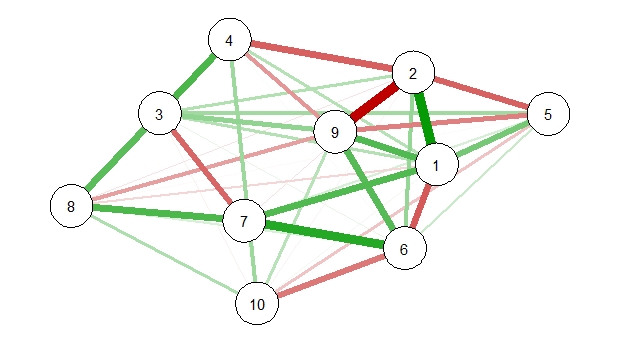}}\hfil
         
     \subfloat[Selected Experts\label{fig:ggm_selection}]{
         \includegraphics[width=0.9\columnwidth]{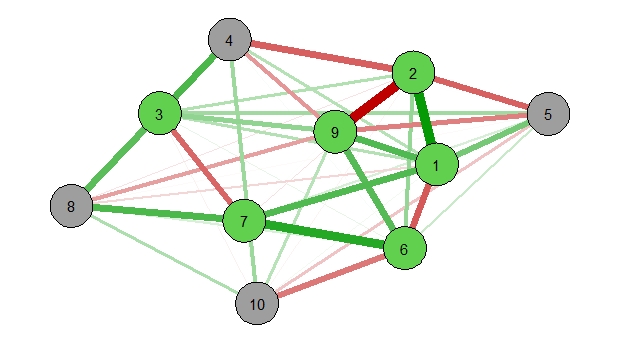}
     }
    \caption{\textbf{Expert Selection} using GGM for a set of 10 local experts from the \textit{Concrete} data set: (a) the experts' graph and (b) selected experts based on $60\%$ of most important experts (green nodes).}
    \label{fig:experts_graph}
\vskip -0.1in
\end{figure}

Figure \ref{fig:experts_graph} depicts the related GGM of local experts' predictions. The \textit{Concrete} data set in Example \ref{example_1} is divided into 10 partitions, i.e., one for each expert, and the experts' predictions are used to quantify interactions between experts in this graph. Figure \ref{fig:ggm} presents the original GGM related to this distributed learning case. Figure \ref{fig:ggm_selection} shows an expert selection scenario when only $60\%$ of the experts are selected based on their importance. In~\cite{jalali2021_arxiv,jalali2021}, the authors studied this selection method in CI-based baselines and reported remarkable improvements over state-of-the-art aggregation approaches in terms of prediction quality.

\begin{algorithm}[!t]
  \caption{Aggregating Dependent Experts Using GGM}\label{alg:gmm}
  \begin{algorithmic}[1]
    \Require Local predictions of GP experts $\mu^*$, sparsity hyperparameter $\lambda$, selection percentage $\alpha$.
	\Ensure Aggregated estimator $y^*_A$.
    \State Calculate sample covariance S of experts' predictions.
    \State Estimate $\hat{\Omega}$ using \textit{GLasso} \eqref{eq:glasso}.
	\State Calculate the importance values $\left\{\mathcal{I}_{1},\mathcal{I}_{2}, \ldots, \mathcal{I}_{M} \right\}$
	\State Sort the importance values to find $\mathcal{I}$ as defined in Definition \ref{def.importance} .
	\State Select $\alpha$ percent of most important experts.
	\State Create the expert set for Aggregation, $\mathcal{M^G}$.
	\State Aggregate selected experts $\mathcal{M^G}$ using \eqref{eq:npae}.
  \end{algorithmic}
\end{algorithm}

Algorithm \ref{alg:gmm} summarizes the aggregation procedure with GMM-based expert selection. Its primary input is given by the local predictions, meaning that this selection method is employed after individual experts' predictions. Therefore, it does not depend on the entry points. Indeed, based on the Definition \ref{def.importance}, only absolute values of conditional dependencies are used for the importance calculation, i.e., $|\hat{\Omega}|$, indicating that the importance is affected only by the amount of the dependency, and not its direction.

Although GGM-based expert selection provides an interpretable method, it suffers from two significant obstacles. First, it needs GLasso to obtain the precision matrix, and the cost of the GLasso is $\mathcal{O}(M^3)$. Hence, for massive data sets with large M, its application is impractical. Besides, the algorithm is not flexible enough to capture the specific behavior of new data points because it provides a static method that selects a fixed set of experts for all test points. Even if an expert can provide accurate prediction for some part of the data set, it will be excluded from the model if it does not have high interactions with the other experts. 

In the following subsections, we propose a novel approach that estimates the essential (i.e., data-related) experts for each new test point by converting the problem into a multi-label classification. The obtained labels for each test point define the data-point-wise selected experts.

\subsection{Multi-label Classification for Flexible Expert Selection} \label{sec:multi-label}
Assigning experts to new entry points in a distributed learning model can be seen as a classification problem. Let's assume that each expert is a class of estimators. The selection problem then for each test point $x^*$ is defined as a multi-label classification task where each instance can be associated with some classes. The main advantage of this method is its flexibility because the selected experts depend on the given test point, and thus different experts can be assigned to different test points.

Assume $x^*$ is a new test point and $\mathcal{E}=\{\mathcal{E}_1,\ldots,\mathcal{E}_M\}$ is the Gaussian experts set, and $\mathcal{L}=\{1,\ldots,M\}$ is the label set. The task is to find $\mathcal{M^C}(x^*)=\{\mathcal{M^C}_1(x^*),\ldots,\mathcal{M^C}_{K}(x^*)\}$ that represents $K$ selected experts to predict at $x^*$. We adapt two prominent classification models to solve this multi-label task without requiring problem transformations, K-nearest neighbors (KNN) and conventional deep neural networks (DNN).

\begin{example} [\textbf{Expert Selection Models}] \label{example_2}
Let's consider an example with five local experts $\mathcal{E}=\{\mathcal{E}_1,\ldots,\mathcal{E}_5\}$ that predict at 10 test points and $K=3$. Figure \ref{fig:experts_selection_scheme} describes the difference between static and dynamic expert selection models in an example with synthetic data points. Figure \ref{fig:cgp_a} depicts the original aggregation where all experts are used, e.g., for the ensemble model in \eqref{eq:npae}. The GGM-based expert selection in \ref{fig:cgp_b} proposes a fixed set of 3 experts $\{ \mathcal{E}_2,\mathcal{E}_4,\mathcal{E}_5\}$ for all new (i.e., 10) entry points even though they do not provide appropriate predictions in some of this 10 test points. The flexibility of the entry-based selection model is depicted in \ref{fig:cgp_c}, where the model assigns different experts to each test point $x^*$ and uses the ability of experts in a better way. 
\begin{figure}[t!]
     \centering
     \subfloat[$\mathcal{E}$ \label{fig:cgp_a}]{
         \includegraphics[width=0.63\columnwidth]{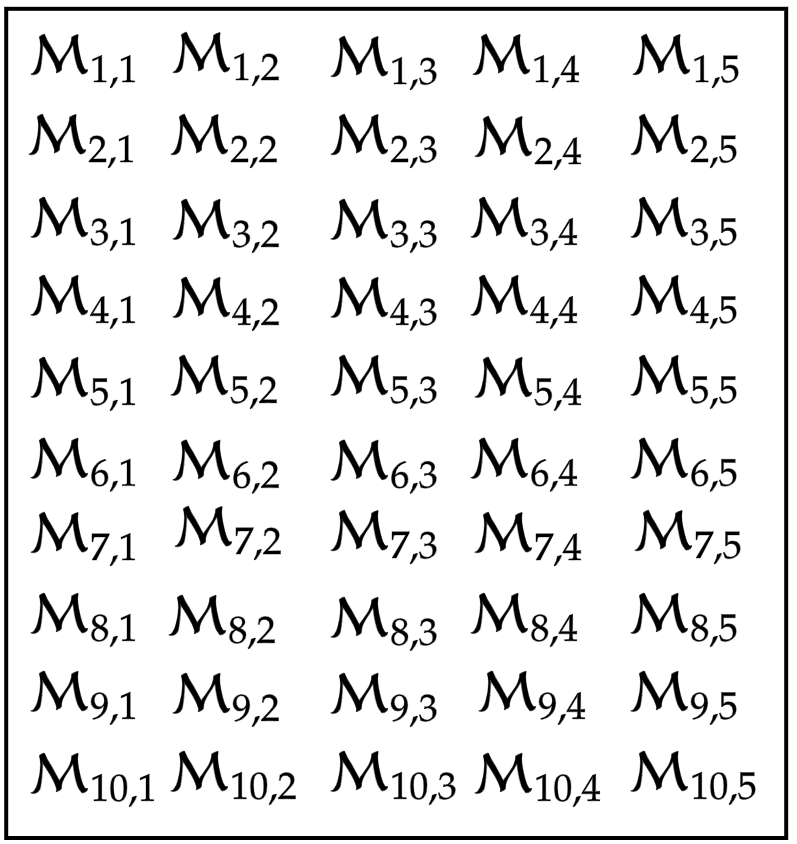}}
         
     \subfloat[$\mathcal{E}^\mathcal{G}$ \label{fig:cgp_b}]{
         \includegraphics[width=0.63\columnwidth]{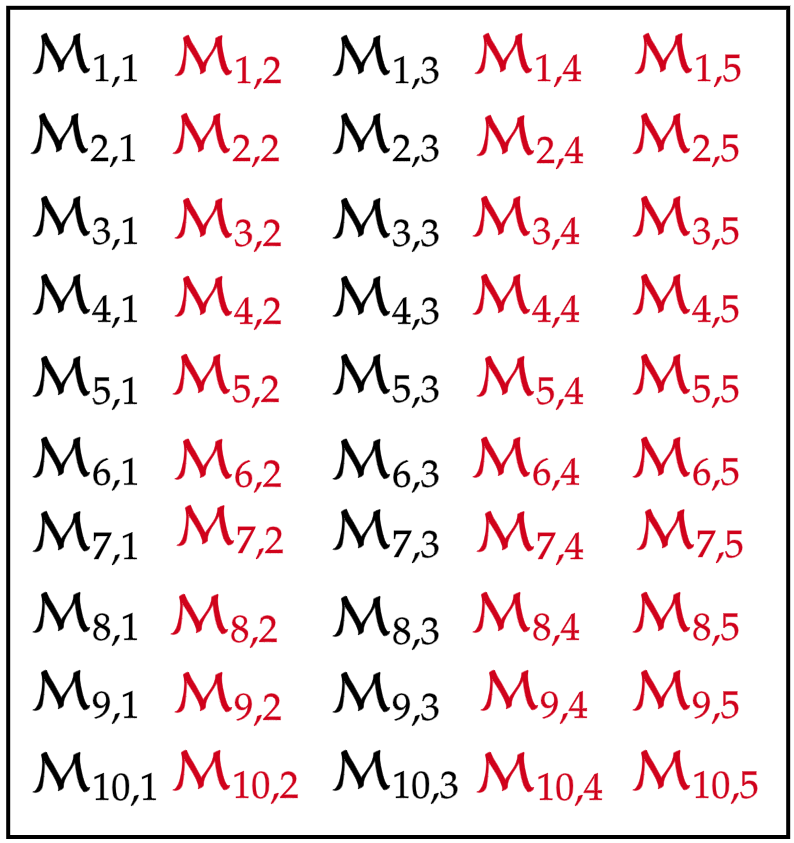}}
         
          \subfloat[$\mathcal{E}^\mathcal{C}$\label{fig:cgp_c}]{
         \includegraphics[width=0.63\columnwidth]{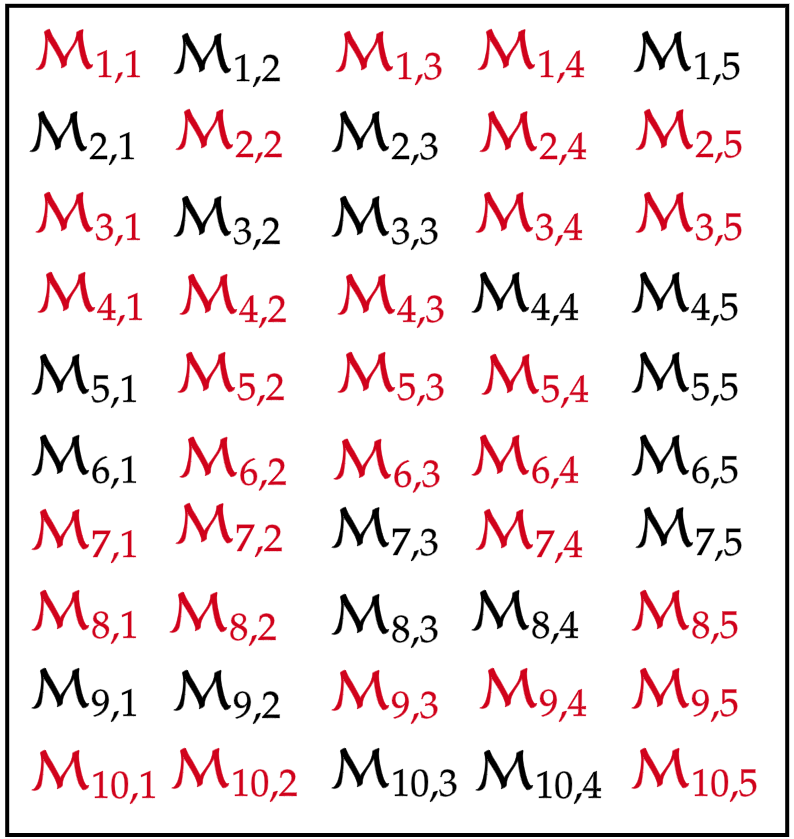}}
    \caption{\textbf{Expert Selection} scheme of both static and entry-dependent models for a setting of 5 experts with 10 test points. Both selection models assign 3 experts to each test point: (a) original set of experts $\mathcal{E}$, (b) static assignment of experts $\mathcal{E}^\mathcal{G}$, and (c) entry-based selection of experts $\mathcal{E}^\mathcal{C}$.}
    \label{fig:experts_selection_scheme}
\vskip -0.1in
\end{figure}
\end{example}

\subsection{K-nearest neighbors (KNN)}

The K-nearest neighbors algorithm \cite{Altman} is a lazy, non-parametric classification approach that uses proximity to classify an individual data point. It is a supervised machine learning algorithm, working off the assumption that similar data points are located near one another. Here, we adopt this algorithm for the experts' assignment in a distributed learning scenario such that the raw training data set is not needed for the selection process and only the partitions' information is used.

Let $\mathcal{P}^{'}= \{\mathcal{P}_1,\ldots,\mathcal{P}_M\}$ be the partitions based on a disjoint partitioning strategy, i.e. K-Means clustering. Also, assume $\mathcal{C}= \{\mathcal{C}_1,\ldots,\mathcal{C}_M\}$ contains the related centroids of the clusters in $\mathcal{P}$. For each test point $x^*$, there is a $1\times M$ vector $dist(x^*, \mathcal{C})$ , in which the $i$'th element is the distance between $x^*$ and $\mathcal{P}_i$, where $dist()$ is a distance metric. Therefore, the adopted KNN algorithm is defined as:
\begin{itemize}
    \item calculate the distance between $x^*$ and the centroids $dist(x^*, \mathcal{C})$
    \item choose $K$ experts with closest centroids to $x^*$
    \item return $\mathcal{M^C}(x^*)$ based on the selected experts.
\end{itemize}

To determine which experts/partitions are closest to a given query point, the distance between the query point and the other data points will need to be calculated using a distance metric. The distance metric helps to form decision boundaries, which partition test points into different subsets/experts. Several distance measures can be chosen, e.g. \textit{Euclidean}, \textit{Manhattan}, \textit{Minkowski}, and \textit{Hamming} distances. In this work, we use the conventional \textit{Euclidean} distance.

The value of $K$ in the KNN algorithm defines how many clusters will be checked to determine the classification of a specific entry point. For example, if $K=1$, the instance will be assigned to the same class as its nearest cluster, which due to discontinuity issues, is not the desired case. Lower values of $K$ can have high variance, but low bias and larger values of $K$ may lead to higher bias, and lower variance \cite{everitt2011}.

\begin{figure}[t!] 
\centering
\subfloat[KNN]{\fbox{\includegraphics[width=\columnwidth]{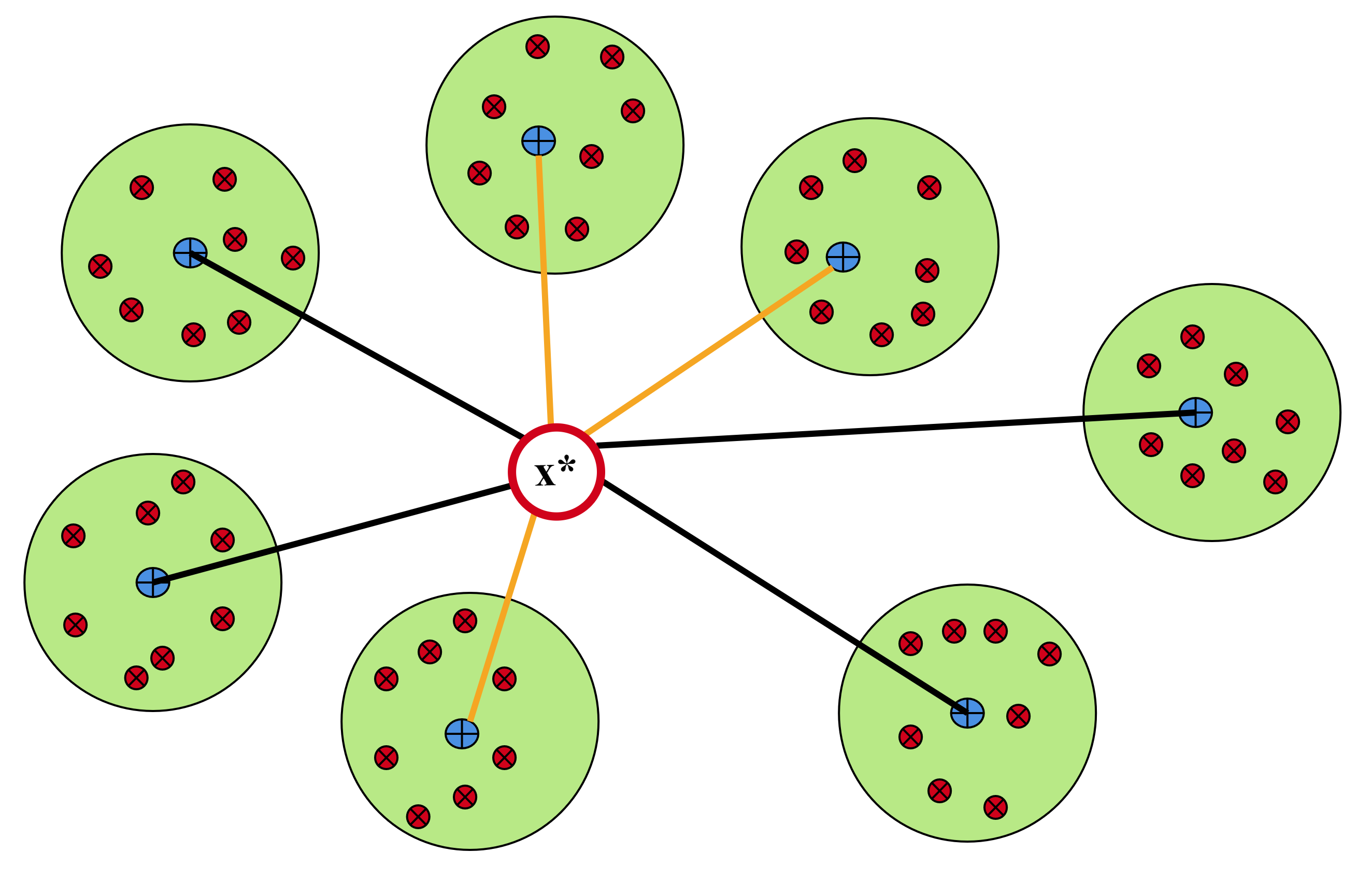}}}\hfil
\caption{Adopted K-nearest neighbors for multi-label classification.}
\label{fig.knn_net}
\end{figure}

Figure \ref{fig.knn_net} schematically shows how the KNN method works for the multi-label classification methods. It represents a KNN framework for a test point $x^*$. The red points are related training points assigned to each partition, and the blue points are the clusters' centroids. The lines between the $x^*$ and the centroids show the distances. The proposed method suggests the orange lines, which are the shortest, and the related experts are assigned to this test point.

\begin{algorithm}[!t]
  \caption{Aggregating Dependent Experts Using KNN}\label{alg:knn}
  \begin{algorithmic}[1]
    \Require Test point $x^* \in X^*$, centroids set $\mathcal{C}$, hyperparameter $K$, Local GPs moments, distance metric.
	\Ensure Aggregated estimator $y^*_A(x^*)$
    \State Calculate distance vector for $x^*$, i.e. $dist(x^*, \mathcal{C})$.
    \State Sort the elements of  $dist(x^*, \mathcal{C})$ ascendingly.
	\State Select the first $K$ experts in the sorted list of expert distances to generate the set of related experts $\mathcal{M^C}(x^*)$.
	\State Estimate local GPs by the experts in $\mathcal{M^C}(x^*)$ using \eqref{eq:local_mean} and \eqref{eq:local_var}.
	\State Aggregate local predictions from Step 4 using \eqref{eq:npae}.
  \end{algorithmic}
\end{algorithm}

Algorithm \ref{alg:knn} summarizes the whole procedure of the KNN-based aggregation. The main advantage of the KNN method is it does not include another training period. The only thing to be calculated is the distance between different points; therefore, it is straightforward to implement and accept new entry data at any time. Besides, instead of $n$ training points, the modified KNN proposed in Algorithm \ref{alg:knn} only uses the $M$ centroids and scales the distance calculations in large data sets. 

Generally, this selection method is defensible and justifiable because Gaussian processes predict better when a test point is close to them. Since the distance-based solution may have some drawbacks in high-dimensional data sets, we propose another multi-label classification solution in the next section that can be effective in high-dimensional cases.

\begin{figure}[!t] 
\centering
\fbox{\includegraphics[width=0.9\columnwidth]{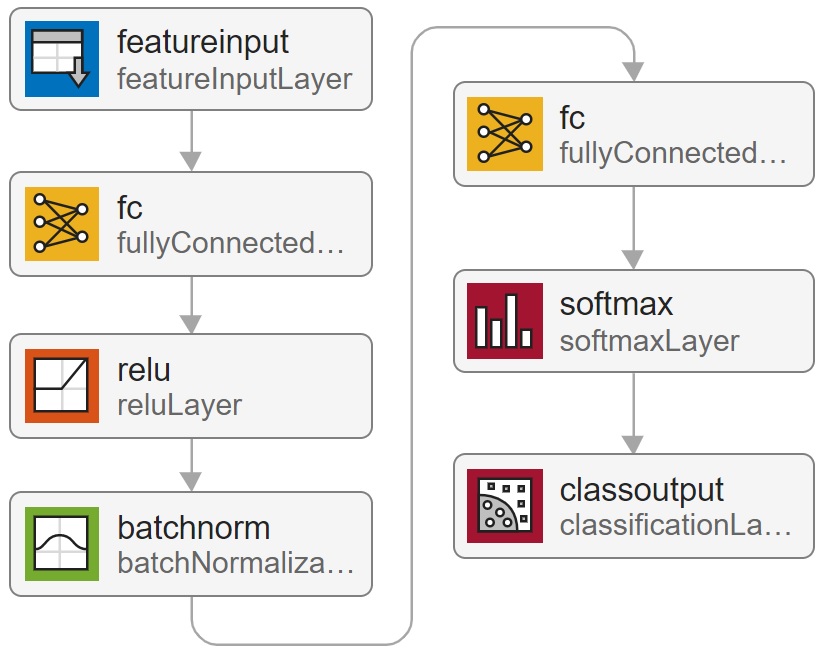}}
\caption{\textbf{Deep Neural Network Architecture} for adopted multi-label classification in DGPs.  }
\label{fig:dnn}
\end{figure}

\subsection{Neural Networks for Classification} \label{sec:dnn}
Conventional deep neural networks (DNNs) are widely used in machine learning problems, especially in classification tasks~\cite{Read2014, Du_2019}. By converting the expert selection task into a multi-label classification task, this supervised learning problem can be solved through DNNs. The capability of the neural networks can compensate for the possible weaknesses of KNN classifiers in dealing with high-dimensional data sets and underlying dependencies between labels.

Multi-label classification can be supported directly by neural networks simply by specifying the number of target labels in the problem as the number of nodes in the output layer. We will define a Multi-layer Perceptron (MLP) model for the multi-label classification task described in subsection \ref{sec:multi-label}. The network requires an input layer that expects $D$ inputs to specify the dimension of $X$, $H$ nodes in the hidden layers, and $M$ nodes in the output layer, indicating the number of experts. Each node in the output layer must use the \textit{softmax} or \textit{sigmoid} activation to predict the label's class membership probability. Finally, the model must fit with the binary cross-entropy loss function and the Adam version of stochastic gradient descent.

To consider the expert selection task as a multi-label classification, a label set $\mathcal{L}=\{1,\ldots,M\}$ contains required classes related to the training data set. For each $x_i \in X$, $i=1,\ldots,N$, the related partition label $l_i\in \mathcal{L}$ is available as an output of the training step in DGP. Therefore, instead of the original training set $(X,y)$, a new set of points and labels $(X,\mathcal{L})$ is constructed for training the DNN. After that, for each test point $x^* \in X^*$ the network will provide a probability vector $P^\mathcal{L}(x^*)$ where $P^\mathcal{L}(x^*)_j$ represents the probability that $x^*$ belongs to the $j$'th expert. The $K$ partitions with highest probabilities in $P^\mathcal{L}(x^*)$ are assigned to $x^*$. 

Figure \ref{fig:dnn} depicts a simple network for a multi-label classification task that assigns local approximation Gaussian process experts to new entry points. 
The network receives training set $X$ as input and their clusters' indices as output. After training the network, test points $x^* \in X^*$ are sent to the neural network, which returns the classifier's raw output values.

\begin{algorithm}[!t]
  \caption{Aggregating Dependent Experts Using DNN}\label{alg:dnn}
  \begin{algorithmic}[1]
    \Require Test point $x^* \in X^*$, training points $X$ and their indices, index set $\mathcal{L}$, hyperparameter $K$, Local GPs moments.
	\Ensure Aggregated estimator $y^*_A(x^*)$
    \State Train the network parameters using $X$ and indices.
    \State Return the classifier’s output values $P^\mathcal{L}(x^*)$, i.e., as produced by the \textit{softmax} layer.
	\State Sort the elements of $P^\mathcal{L}(x^*)$ descendingly.
	\State Select the first $K$ indices of the sorted $P^\mathcal{L}(x^*)$.
	\State Create the expert set for $x^*$, $\mathcal{M^C}(x^*)$.
	\State Estimate local GPs by the experts in $\mathcal{M^C}(x^*)$ using \eqref{eq:local_mean} and \eqref{eq:local_var}.
	\State Aggregate local predictions from Step 6 using \eqref{eq:npae}.
  \end{algorithmic}
\end{algorithm}

\begin{figure}[t!]
     \centering
     \subfloat[SMSE \label{fig:smse_selection}]{
         \includegraphics[width=0.95\columnwidth]{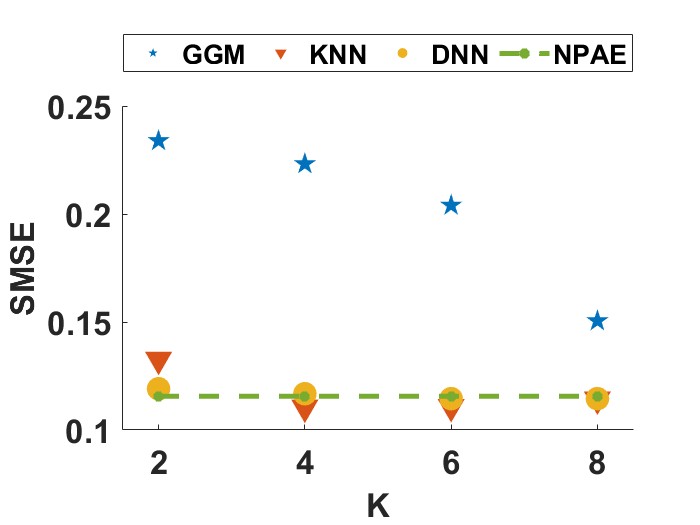}}\hfil
         
     \subfloat[MSLL \label{fig:msll_selection}]{
         \includegraphics[width=0.95\columnwidth]{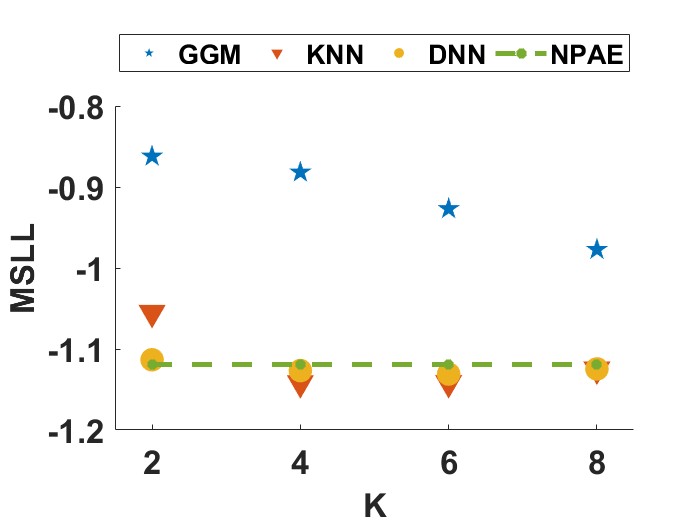}}
    \caption{\textbf{Expert Selection} prediction qualities of different experts selection methods compared to original baseline, NPAE, from \textit{Concrete} data set. }
\label{fig:experts_selection}
\vskip -0.1in
\end{figure}

Figure \ref{fig:experts_selection} depicts the prediction quality of expert selection methods on the \textit{Concrete} data set with 10 partitions for different values of $K$. As can be seen, multi-label-based expert selection models provide higher quality predictions with lower deviation from the NPAE model. The quality of the classification-based aggregations changes less as the selection parameter $K$ increases. The case $K=10$ leads to the original NPAE baseline (the dashed green lines in Figures \ref{fig:smse_selection} and \ref{fig:msll_selection}), and both KNN and DNN return proper error values in both plots.

\section{Discussion} \label{discussion}
This section considers some specific aspects of the expert selection models. 

\subsection{Restrictive Assumptions}
The GGM approach assumes that the nodes in the graph are random, and their joint distribution is Gaussian. This normality assumption leads to a Gaussian likelihood, and GLasso solves this optimization problem. This assumption is strong, and in some cases, it can be restrictive. Various solutions have been proposed to relax this assumption by considering the model as a nonparametric problem and solving after some smooth monotone transformation \cite{Lafferty, Mulgrave_2020, li, Solea_2021} at the cost of requiring higher time complexity. On the other hand, expert selection using multi-label classification does not need any distributional assumption. Therefore it can be used as a general expert selection method in distributed/federated learning models and not only in the context of local approximation of GPs.

Besides, the classification-based expert allocation can also be considered a self-attention mechanism that implicitly captures relationships between data points. Recently, the explicit modeling of self-attention between all data points has been shown to boost the classification performance~\cite{Vaswani2017, Kossen2021}. In our case, to explain the dependencies between training and test points, the expert ensembles do not use the original training points. Instead, the final prediction benefits from the critical information of training data captured by the partitions‘ centroids and the corresponding indices for KNN and DNN, respectively.

\subsection{Computational Costs of Expert Selection Models} \label{section:computational_cost}
The aggregation cost in all three selection methods, i.e., GGM, KNN, and DNN, is $\mathcal{O}(N_t K^3)$ where $K$ is the number of selected experts and $N_t$ is the number of test observations. However, their selection strategies lead to different computational costs. GGM in Algorithm \ref{alg:gmm} needs GLasso to estimate the precision matrix, and its computational cost is $\mathcal{O}(M^3)$, where the $M$ is the number of initial experts and is challenging for large $M$. Indeed, the sparsity parameter can affect the cost such that choosing a smaller value for $\lambda$ leads to a dense graph with a more considerable computational cost. 

The cost of the KNN approach \ref{alg:knn} is obtained by considering the cardinality of the training set, which refers to the number of possible labels that a feature can assume, in our case $M$, the dimension of each sample, i.e., $D$, and also the hyperparameter $K$. The computation time for calculating the distances is usually negligible compared to the rest of the algorithm. However, we consider this aspect as well in the overall cost estimation. Algorithm \ref{alg:knn} computes the distance between the new observation and each centroid point, requiring $\mathcal{O}\left(MD\right)$ work for an iteration and therefore $\mathcal{O}\left(KMD\right)$  work overall to select $K$ closest centroids.

The cost of the DNN approach in Algorithm \ref{alg:dnn} depends on the network structure, i.e., the number of layers $L$, the input dimension $D$, the output dimension $M$, and the number of hidden units. Let $U_i$ represent the number of units in the $i$'th layer ($i=1,\ldots, L$), where $U_1$ and  $U_L$ represent the number of units in the input and output layers, respectively. The computational complexity is thus $\mathcal{O} \left( N (U_1U_2 + \ldots + U_{L-1}U_L) \right)$. 

In conclusion, both methods described in Algorithms \ref{alg:knn} and \ref{alg:dnn} have linear complexity concerning $M$. Therefore, they are more efficient when the number of partitions is an enormous value, unlike the cubical dependency in Algorithm \ref{alg:gmm}.

\subsection{Activation Functions in DNN}

Using a \textit{softmax} output layer for the DNN-based classification in Section \ref{sec:dnn} leads to a probability vector $P^\mathcal{L}(x^*)$ of the output values. 
Hence, when the probability of one class increases, the probability of at least one of the other classes has to decrease by an equivalent amount. Since the labels represent the interdependent experts, using the \textit{softmax} function for the classification layer is reasonable.

\begin{figure}[t!]
     \centering
     \subfloat[SMSE \label{fig:smse_dnn_activation}]{
         \includegraphics[width=0.95\columnwidth]{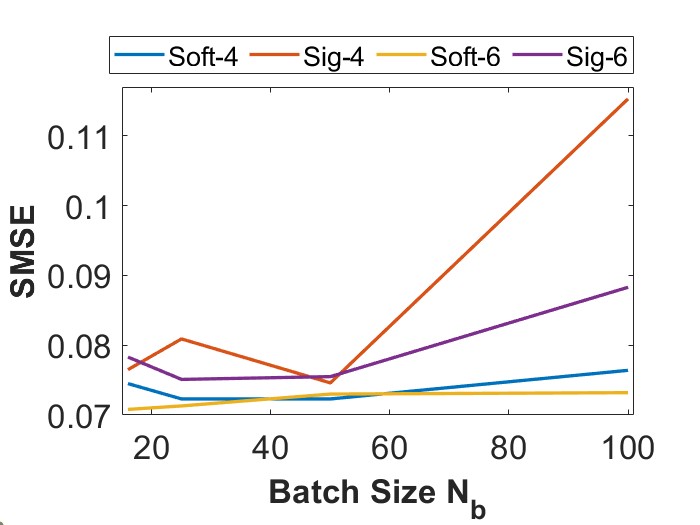}}\hfil
         
     \subfloat[MSLL \label{fig:msll_dnn_activation}]{
         \includegraphics[width=0.95\columnwidth]{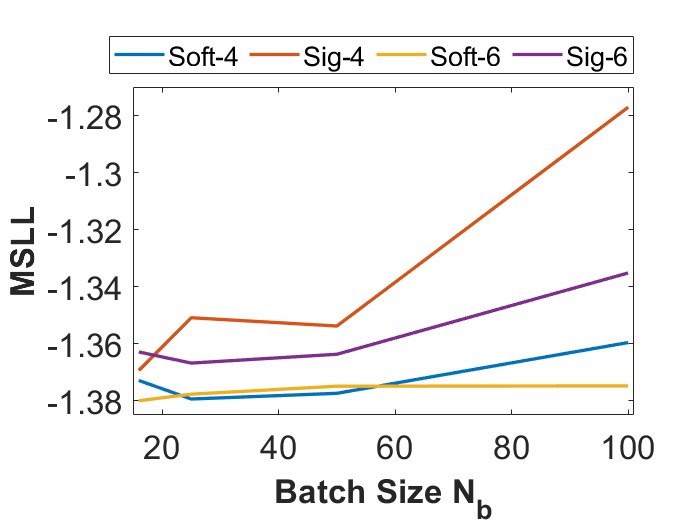}}
    \caption{\textbf{Activation Functions} for the final (i.e., output) layer of the DNN classifier: prediction quality for $K=4$ and $K=6$ in an experiment from the \textit{Concrete} data set with 10 experts.}
\label{fig:activation_function}
\vskip -0.1in
\end{figure}

Figure \ref{fig:activation_function} explains how different activation functions can affect the prediction quality. It considers the \textit{Concrete} data set with $M=10$ and different mini-batch sizes $N_b=\{16,25,50,100\}$. Both activation functions \textit{softmax} and \textit{sigmoid} have been used to select $K=4$ and $K=6$ experts. The quality of the related aggregations confirms that due to the interaction between labels, the \textit{softmax} activation leads to much better results, and its sensitivity to the size of the mini-batches is lower than for the \textit{sigmoid} activation.

\subsection{Expert Selection for CI-Based Baselines}

In~\cite{jalali2021}, the authors extended the GGM-based expert selection for CI-based ensembles. They introduce an extra step aiming to exclude the unimportant experts from the model before using the weight parameters $\beta$. The same procedure can be used for entry-based expert selection methods. In this case,  for a test point $x^*$ in \eqref{eq:CI_ensemble}, only $K$ experts are used, and the selection is based on the Algorithm \ref{alg:knn} or Algorithm \ref{alg:dnn}. Since these models are fast, the selection parameter $K$ can be also set to relatively large values. 

\begin{table}[!ht]
  \centering
  \caption{Expert Selection in CI-Based Baselines.}\label{tab:CI_knn}
 \scalebox{1.15}{  \begin{tabular}{ccccc}
	\hline
	Model & Expert Selection & SMSE  & MSLL & Time (s)  \\\hline
	\multirow{2}*{GPoE} & - & 0.138 & -0.876 & 0.03 \\
	& KNN & \textcolor{red}{0.115} & \textcolor{red}{-0.916} & \textcolor{red}{0.03}   \\ \hline
	\multirow{2}*{RBCM} & - & 0.0993 & 0.396 & 0.03\\		
		& KNN & \textcolor{red}{0.091} &\textcolor{red}{0.156} & \textcolor{red}{0.03} \\ \hline
	\multirow{2}*{GRBCM} & - & 0.1093 & -1.103 & 0.06\\		
		& KNN & \textcolor{red}{0.089} &\textcolor{red}{-1.21} & \textcolor{red}{0.06} \\ 
	\hline
  \end{tabular}}
\end{table}

Table \ref{tab:CI_knn} describes the effect of the selection scenario on CI-based ensembles using KNN on the \textit{Concrete} data set with $M=10$ and $K=6$. Although this modification can not improve the asymptotic properties of the baselines, it raises their prediction quality. At the same time, the running times of both original and modified models are indistinguishable.

\section{Experiments} \label{experiments} 
The quality of the expert selection methods is assessed in this Section. We consider the prediction quality and the required prediction time of the proposed and state-of-the-art distributed GP models using both simulated and real data sets. The quality of predictions is evaluated by the standardized mean squared error (SMSE) and the mean standardized log loss (MSLL). The standard squared exponential kernel with automatic relevance determination and a Gaussian likelihood is used. Since the disjoint partitioning of training data captures the local features more accurately and outperforms random partitioning \cite{Liu2018,jalali2021}, it is mainly used in our experiments. The sparsity parameter in the GGM-based expert selection method is set to $\lambda=0.1$, \textit{Euclidean} norm measures the distances in KNN, and a neural network with a single hidden layer is used for the DNN classification. The experiments have been conducted in MATLAB using the GPML package\footnote{\url{http://www.gaussianprocess.org/gpml/code/matlab/doc/}}. 

\begin{figure*}[t!]
     \centering
     \subfloat[NPAE \label{fig:npae_CI}]{
         \includegraphics[width=0.9\columnwidth]{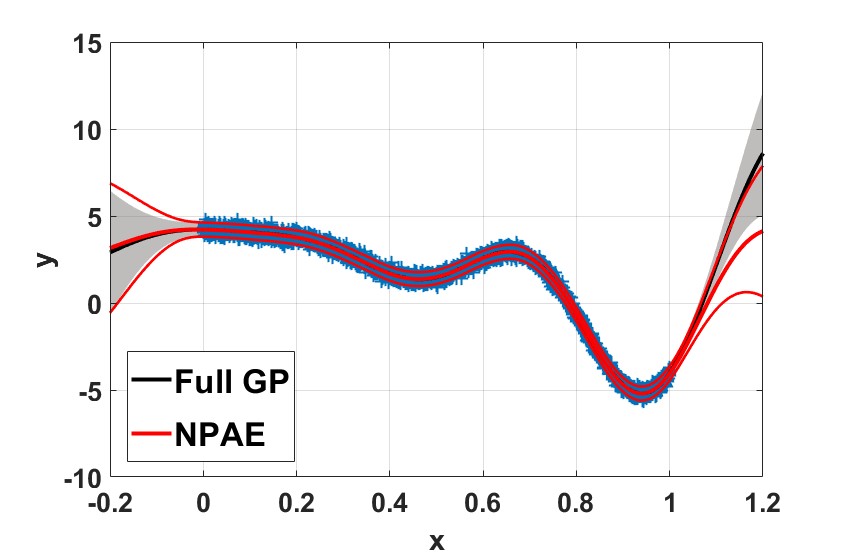}}
     \subfloat[GGM \label{fig:ggm_CI}]{
         \includegraphics[width=0.9\columnwidth]{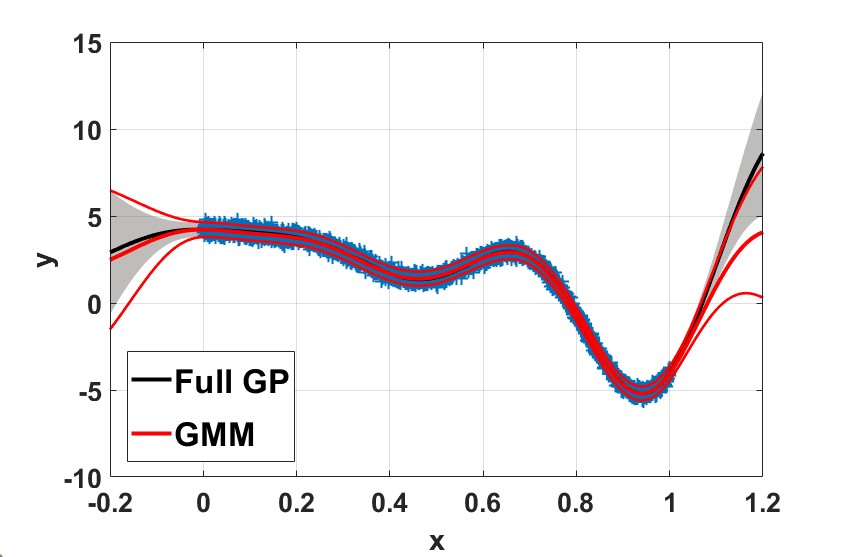}}
         
     \subfloat[KNN \label{fig:knn_CI}]{
         \includegraphics[width=0.9\columnwidth]{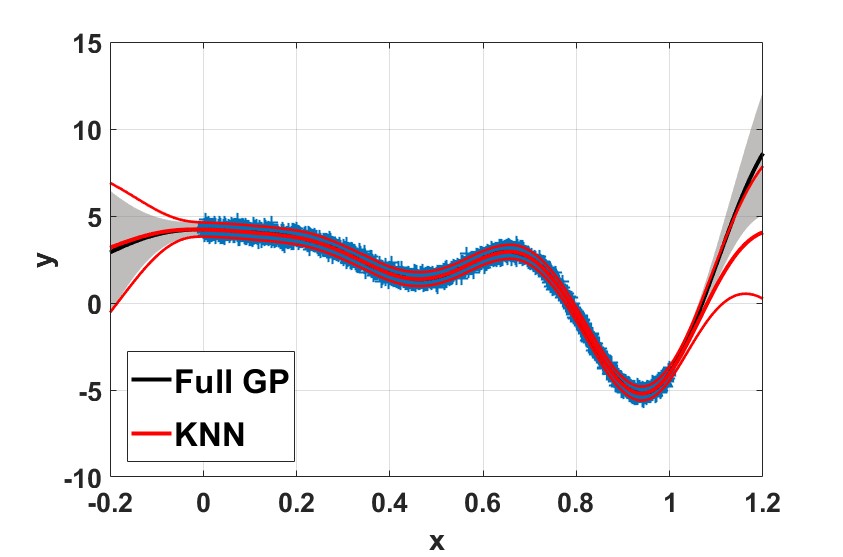}}
     \subfloat[DNN \label{fig:dnn_CI}]{
         \includegraphics[width=0.9\columnwidth]{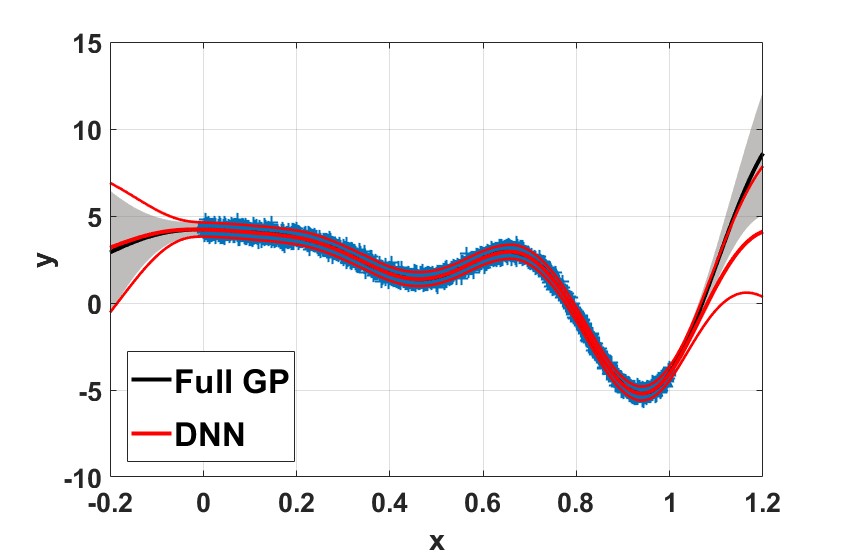}}   
\caption{\textbf{99\% Confidence Interval} of NPAE, expert selection methods and full GP for $n=3\times 10^3$ training points from Equation \eqref{f_x} and $M=10$ (partition size $m_0=300$) with $K$-means partitioning. The GGM, KNN, and DNN results are based on $K=5$ selected experts. We see that the DNN and the KNN approximations to the full GP are practically indistinguishable from the NPAE approximation; all three techniques are quality-wise superior to the GGM approach.}
\label{fig:confidence_intervals}
\vskip -0.1in
\end{figure*}

\subsection{Sensitivity Analysis} \label{sec.5.1}
In this section, we investigate the influence of hyperparameters on the prediction quality and computational cost of the proposed methods and available baselines. First, we consider the aggregations of dependent experts with the selection step using a synthetic one-dimensional data set. Then, we use a medium-scale real-world data set to study how hyperparameters affect the results in a complex multi-dimensional data set.

\subsubsection{Synthetic Example} \label{sec.5.1.1}

The first experiment evaluates the effect of hyperparameters $M$ and $K$ on prediction quality and computation time in different selection scenarios. It is based on simulated data of a one-dimensional analytical function \cite{Liu2018},
\begin{equation}
f(x) = 5x^2sin(12x) + (x^3 -0.5)sin(3x-0.5)+4cos(2x) + \epsilon,  \label{f_x}
\end{equation}
where $\epsilon \sim \mathcal{N}\left(0, (0.2)^2\right)$. We generate $n$ training points in $[0,1]$, and $N_t=0.1n$ test points in $[-0.2,1.2]$.  The data is normalized to zero mean and unit variance. We vary the number of experts to consider different partition sizes. The $K$-means method is used for the partitioning to compare the prediction quality of the proposed selection methods with other baselines. Since the quality of CI-based methods is low, they are excluded from these experiments. 

Figure \ref{fig:confidence_intervals} depicts the 99\% confidence interval of NPAE, expert selection-based aggregations, and the full Gaussian process. In the experiment, $n=3 \times 10^3$ training data points from Equation \eqref{f_x} are used. The training set is divided into $M=10$ partitions, i.e., partition size $m_0=300$, with K-means clustering, and $K=5$ agents are used for the final prediction. The confidence intervals of KNN and DNN are closer to the original baseline NPAE, and their predictions (mean of the predictive distribution) are close to the full GP. For test points out of the training set domain, the multi-label classification leads to accurate expert selection results; see for example the interval $[-0.2,0]$ in the GGM plot, which shows a significant deviation from the standard GP.

\begin{figure*}[t!]
     \centering
     \subfloat[SMSE and $n=3000$ \label{fig:smse_K}]{
         \includegraphics[width=0.66\columnwidth]{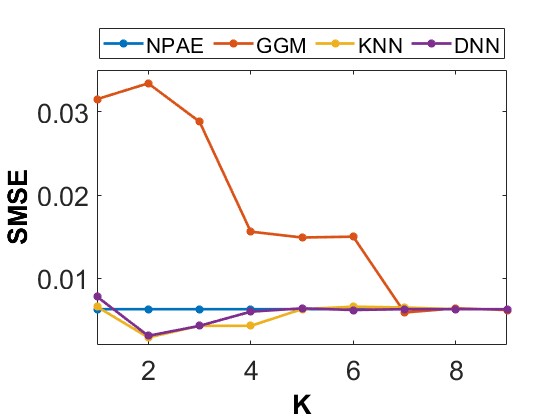}}
     \subfloat[MSLL and $n=3000$ \label{fig:msll_K}]{
         \includegraphics[width=0.66\columnwidth]{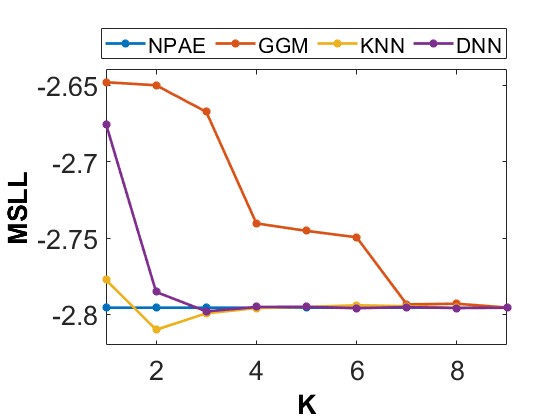}}
     \subfloat[Time and $n=3000$ \label{fig:time_K}]{
         \includegraphics[width=0.66\columnwidth]{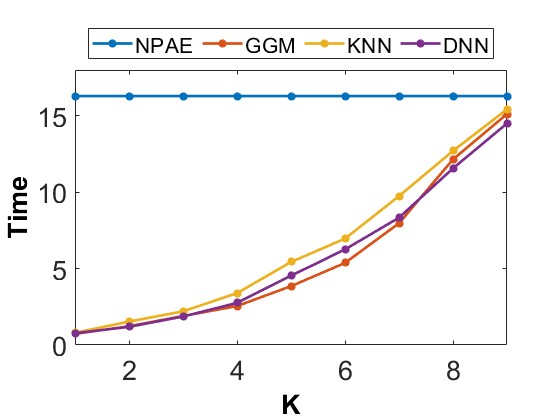}}
\caption{\textbf{Prediction quality} of available baselines as a function of experts for $3\times10^3$ training points and partition size $m_0=300$ with $K$-means partitioning.}
\label{fig:fx_K}
\vskip -0.1in
\end{figure*}

\begin{figure*}[t!]
     \centering
     \subfloat[SMSE and $n=3000$ \label{fig:smse_3000}]{
         \includegraphics[width=0.66\columnwidth]{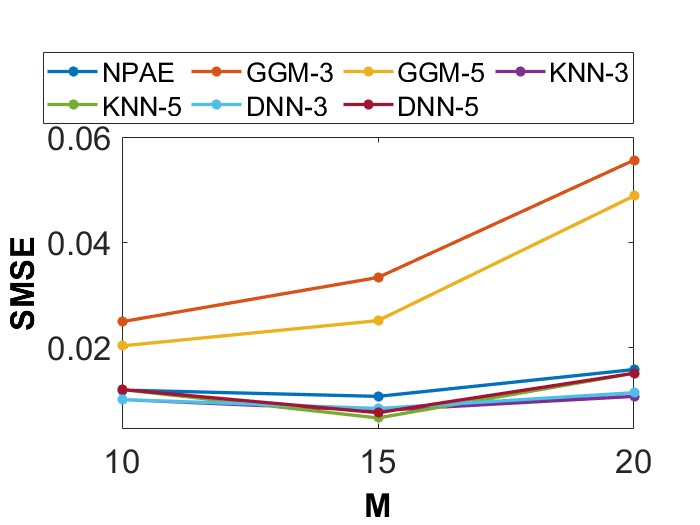}}\hfil
     \subfloat[MSLL and $n=3000$ \label{fig:msll_3000}]{
         \includegraphics[width=0.66\columnwidth]{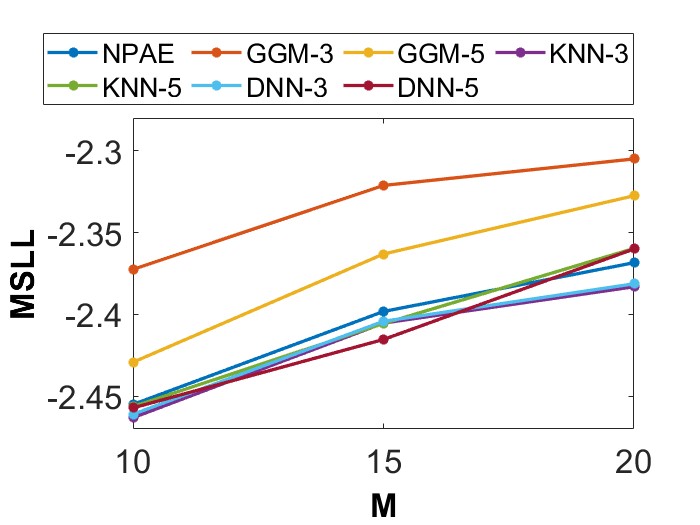}}
     \subfloat[Time and $n=3000$ \label{fig:time_3000}]{
         \includegraphics[width=0.66\columnwidth]{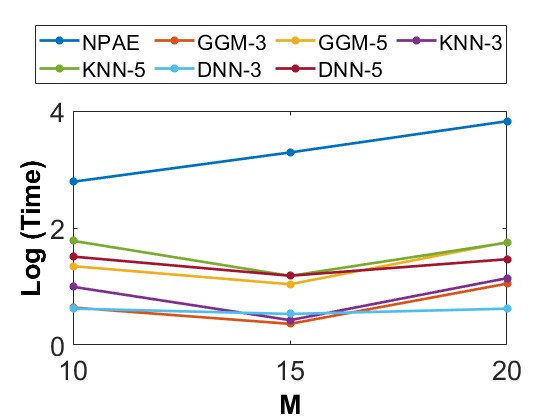}}
    
    \subfloat[SMSE and $n=5000$ \label{fig:smse_5000}]{
     \includegraphics[width=0.66\columnwidth]{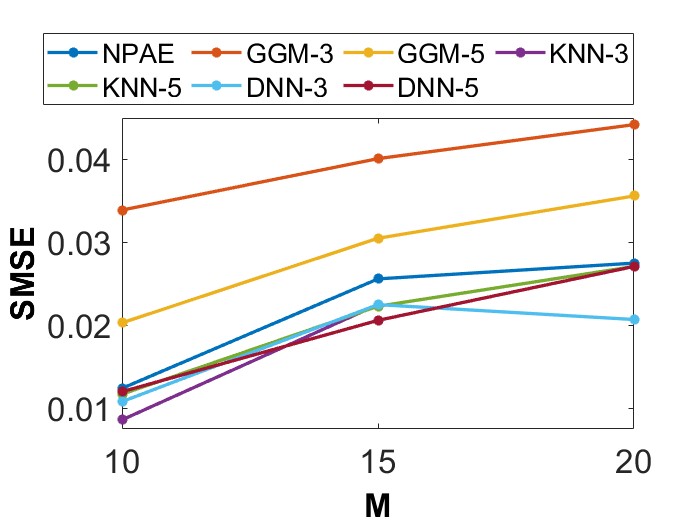}}\hfil
    \subfloat[MSLL and $n=5000$ \label{fig:msll_5000}]{
     \includegraphics[width=0.66\columnwidth]{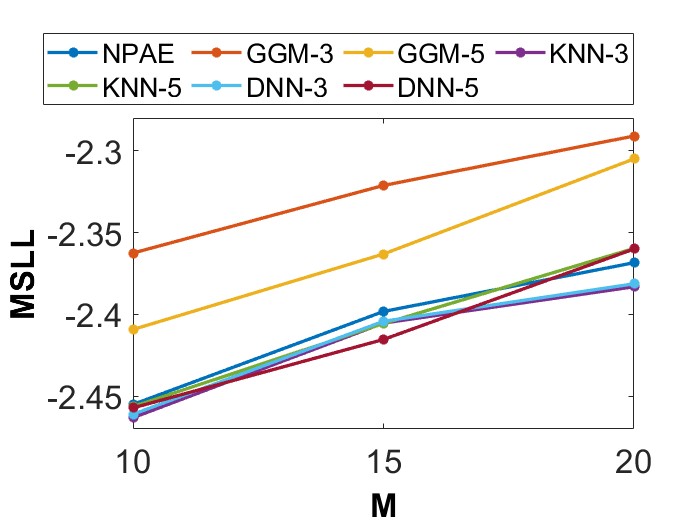}}
    \subfloat[Time and $n=5000$ \label{fig:time_5000}]{
     \includegraphics[width=0.66\columnwidth]{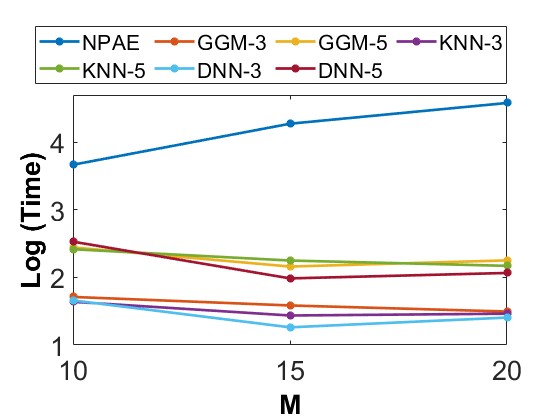}}
         
\caption{\textbf{Prediction quality} and running time of DGP baselines for $3000$ and $5000$ training points from Equation \eqref{f_x} with different numbers of partitions, $M=\{10,15,20\}$. For expert selection-based methods, $K=\{3,5\}$ experts have been selected for the final aggregation.}
\label{fig:fx_prediction_quality}
\vskip 0.1in
\end{figure*}

Figure \ref{fig:fx_K} depicts the prediction quality of expert selection methods compared to NPAE for $3\times10^3$ training points and partition size $m_0=300$ (i.e., ten experts) with $K$-means partitioning.  The x-axis shows the number of selected experts ($K$) used in the final aggregation. We vary the number of selected experts for selection methods and use the NPAE as a baseline that refers to $K=10$. The plots in \ref{fig:smse_K} and \ref{fig:msll_K} indicate that multi-label classification leads to a selection strategy with fast convergence to the original estimator. 

When the number of experts increases, the prediction errors of KNN and DNN do not show significant changes, which is an indication of predictive stability. On the other hand, GGM needs more experts to provide closer results to NPAE and has a slow convergence procedure. For $K\geq 7$, the error values of all three methods are almost the same. Figure \ref{fig:time_K} indicates that the computational costs of the aggregations based on GGM, KNN, and DNN are approximately the same. 

We evaluate the prediction quality of the expert selection methods using $n=3\times 10^3$ and $n=5\times 10^3$ training points and different numbers of experts, $M=\{10,15,20\}$. All related baselines are used in this experiment with $K=\{3,5\}$ selected experts. Figure \ref{fig:fx_prediction_quality} depicts the results of both generated samples. KNN and DNN aggregations in both samples have remarkable prediction qualities, and their SMSE and MSLL values are close to each other, which means both classification methods return almost similar results.

\begin{table*}[hbt!]
\caption{SMSE and MSLL of different baselines on the \textit{Pumadyn} data set for different number of partitions strategies. Both dependent (D) and conditionally independent (CI) aggregation methods are used.}
\label{table.pumadyn}
\vskip -0.3in
\begin{center}
\begin{small}
\begin{sc}
\scalebox{1.2}{\begin{tabular}{lccccccr}
\toprule
\multicolumn{1}{c}{} &
\multirow{2}{*}{} &
\multicolumn{2}{c}{$M=10$} &
\multicolumn{2}{c}{$M=15$} &
\multicolumn{2}{c}{$M=20$} \\
\toprule
 Model &Type & SMSE  & MSLL  & SMSE & MSLL & SMSE & MSLL  \\
\midrule
GPoE    &CI& 0.0487 & -1.5092& 0.0489& -1.5087& 0.0501& -1.4815  \\
GRBCM      &CI& 0.049& -1.5129 & 0.0490 & -1.5083& 0.0486& -1.5133  \\
\midrule
NPAE    &D& \textbf{0.0462} & \textbf{-1.5397} & \textbf{0.0473} &\textbf{-1.5271}& \textbf{0.0470} &\textbf{-1.5285}\\
\midrule
\textcolor{blue}{GMM-5} & D & 0.0477 & -1.5249& 0.0485& -1.5103 & 0.0481& -1.5180  \\
\textcolor{blue}{GGM-7} &D & 0.0471 & -1.5307 & 0.0481& -1.5165 & 0.0478& -1.5208 \\
\textcolor{blue}{KNN-5} &D & 0.0467 &-1.5364& 0.0477 &-1.5236 & 0.0475 & -1.5234    \\
\textcolor{blue}{KNN-7} &D & \textbf{0.0462} &\textbf{-1.5402} & \textbf{0.0474} &\textbf{ -1.5266} & \textbf{0.0470} &\textbf{ -1.5285}  \\
\textcolor{blue}{DNN-5} &D& 0.0465 & -1.5370 & 0.0477 &-1.5232& 0.0476 &-1.5216\\
\textcolor{blue}{DNN-7} &D& \textbf{0.0460} & \textbf{-1.5418} & \textbf{0.0475} & \textbf{-1.5253} & \textbf{0.0470} &\textbf{-1.5285}\\
\bottomrule
\end{tabular}}
\end{sc}
\end{small}
\end{center}
\vskip 0.1in
\end{table*}

\begin{figure*}[t!]
     \centering
     \subfloat[\normalsize Aggregation with 50\% of Experts \label{fig:pumadyN_time_5}]{
         \includegraphics[width=0.99\columnwidth]{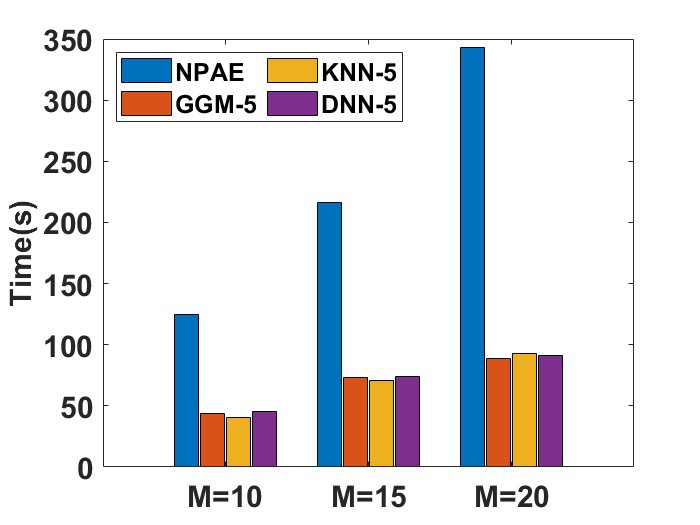}}\hfil
     \subfloat[\normalsize Aggregation with 70\% of Experts\label{fig:pumadyN_time_7}]{
         \includegraphics[width=0.99\columnwidth]{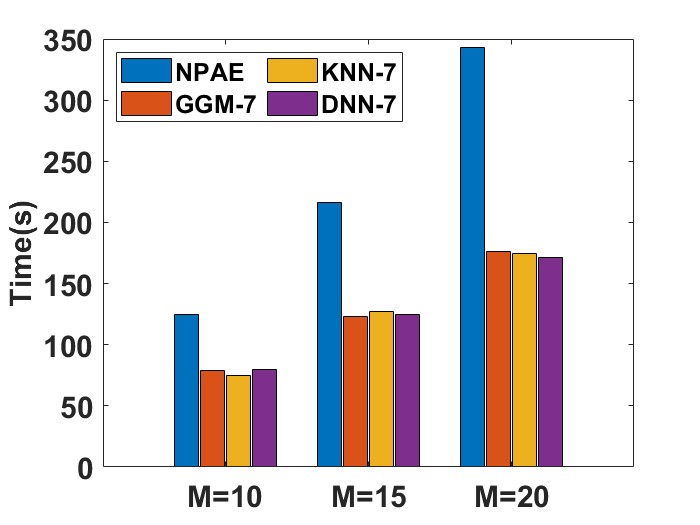}}
    
\caption{\textbf{Prediction Time} (seconds) of different aggregation for disjoint partitioning in \textit{Pumadyn} data set. The training data set is divided into 10, 15, and 20 subsets and a 50\% and 70\% of experts are selected for final aggregation.}
\label{fig.pumadyN_time}
\vskip -0.1in
\end{figure*}

On the other hand, when the ratio of the selected experts to the total number of experts $K_M=\frac{K}{M}$ decreases, the prediction quality of GGM-based models decreases drastically. For instance in Figure \ref{fig:smse_3000}, the differences between SMSE values of \textit{GGM-3} and \textit{GGM-5} at $M=20$ are almost twice the SMSE at $M=15$. Indeed, GGM requires more experts to provide qualitative predictions, and the difference between the SMSE and MSLL of \textit{GGM-3} and \textit{GGM-5} indicates this fact. At the same time, the quality of KNN and DNN does not change significantly when $K$ increases from 3 to 5.

Figures \ref{fig:time_3000} and \ref{fig:time_5000} show the running time of the baselines that consider the dependencies between experts. All expert selection-based aggregations share the same prediction process (of $\mathcal{O}(N_t K^3)$), differing only in the selection task. Besides enhancing the number of selected experts, K increases the computational cost of selection methods because it raises the prediction cost $\mathcal{O}(N_t K^3)$, see Section \ref{section:computational_cost}. In these experiments with smooth 1D data points, the running times of the GGM, the KNN, and the DNN are of the same rate.   

\subsubsection{Multi-Dimensional Real-World Data Set} \label{sec.5.1.2}

The relative number of experts $K_M=\frac{K}{M}$ defined in Section \ref{sec.5.1.1} indicates the percentage of the initial experts selected to be used in the final predictive distribution. In this section, we use a medium-scale real-world data set and $K_M$ to appraise the efficacy of the data assignment strategy on the prediction quality. \textit{Pumadyn}\footnote{\url{https://www.cs.toronto.edu/~delve/data/pumadyn/desc.html}} is a generated data set with 32 dimensions and 7,168 training points and 1,024 test points. The disjoint partitioning divides the data set into 10, 15, and 20 subsets. We consider the GPoE\cite{Deisenroth}, GRBCM\cite{Liu2018}, NPAE\cite{Rulliere}, GGM-based aggregation \cite{jalali2021}, and proposed classification-based methods with K-means clustering. The penalty term $\lambda$ is 0.1 for GGM, and a neural network with a single hidden layer and 50 hidden units is used in this experiment.

The expert selection approaches based on GGM, KNN, and DNN, use $K_M=0.5$ and $K_M=0.7$, which means $50\%$ and $70\%$ of available experts are selected, respectively. Table \ref{table.pumadyn} depicts the prediction quality of local approximation methods for the \textit{Pumadyn} data set. The column Type shows the interactions between experts in the aggregation method, D for dependent experts, and CI for conditionally independent experts. The \textit{GGM-5}, \textit{GGM-7}, \textit{KNN-5}, \textit{KNN-7}, \textit{DNN-5}, \textit{DNN-7}, and NPAE are the dependency-based methods while \textit{GPoE} and \textit{GRBCM} are CI-based aggregations. The numbers after the names of the methods indicate the ratio. For instance, \textit{KNN-5} and \textit{KNN-7} refer to KNN with $K_M=0.5$ and $K_M=0.7$, respectively.

NPAE is a basis for comparison because it is the best linear unbiased predictor (BLUP). The multi-label classification methods provide accurate results, and their derivatives with $K_M=0.5$ and $K_M=0.7$ are close to the NPAE. This performance shows the fact that convergence occurs faster in these methods. However, the proficiency of the GGM method is sensitive to the number of agents and has more deviation from the NPAE when $K_M$ is small. Both KNN and DNN with $50\%$ of the experts return appropriate approximations. They offer a significant improvement in prediction quality when $K_M=0.7$, and for $M=10$, they outperform the BLUP baseline, i.e., the NPAE method. It happens because the selection step properly excludes only weak experts at each test point.

Figure \ref{fig.pumadyN_time} depicts the running time of different aggregations with dependent experts for \textit{Pumadyn} data set. The training data set is divided into $M=\{10,15,20\}$ partitions, and the prediction time of the related baselines is compared in two different cases, with 50\% and 70\% of original experts. In both cases, the NPAE method is used as a baseline that uses all dependent experts. Based on the plots, the prediction time of GGM, KNN, and DNN are almost at the same rate. However, Table \ref{table.pumadyn} shows that GGM can not provide competitive prediction quality compared to the other selection methods. For instance, the SMSE and MSLL values of \textit{DNN-5} for all values of $M$ are lower than those of \textit{GGM-5} and \textit{GGM-7}. This issue is also confirmed by Figures \ref{fig:fx_K} and \ref{fig:fx_prediction_quality}. The main reason for this issue is the low convergence rate of the GGM, which requires more experts.

\subsection{Prediction Quality in Real-World Data Sets} \label{sec.5.2}

In this section, we use real-world data sets to evaluate the prediction quality of proposed aggregations and compare them with available baselines. The baselines that we use here are GPoE with uniform weights\cite{Deisenroth}, RBCM\cite{Deisenroth}, GRBCM\cite{Liu2018}, NPAE\cite{Rulliere}, GGM-based aggregation\cite{jalali2021}, KNN and DNN based ensemble methods. Various real-world data sets with different sizes and dimensions are used here which have been explained in Table \ref{tab:datasets}.

\begin{table}[!ht]
  \centering
  \caption{Real-World Data Sets.}\label{tab:datasets}
 \scalebox{1.08}{ \begin{tabular}{ccccc}
	\hline
	Data Set & ($\#$) Observations & n & $N_t$ &  D   \\\hline
	
    \textit{Airfoil}\tablefootnote{\url{https://archive.ics.uci.edu/ml/datasets/airfoil+self-noise}} & 1,503 & 1,203 & 300 &  5 \\ \hline
    
	\textit{Parkinson}\tablefootnote{\url{https://archive.ics.uci.edu/ml/datasets/parkinsons+telemonitoring}} & 5,875 & 5,000 & 875 &  20  \\ \hline    
	
	\textit{Pole Telecom}\cite{Nguyen} & 15,000 & 10,000 & 5,000 & 26\\ \hline		
	
	
	\textit{Protein}\tablefootnote{\url{https://archive.ics.uci.edu/ml/datasets/Physicochemical+Properties+of+Protein+Tertiary+Structure}} & 45,730 & 40,000 & 5,730 & 9 \\ \hline		
	
	\textit{Sacros}\tablefootnote{\url{http://www.gaussianprocess.org/gpml/data/}} & 48,938 & 44,489 & 4,449 & 21  \\ \hline
	
	\textit{Song}\tablefootnote{\url{https://archive.ics.uci.edu/ml/datasets/yearpredictionmsd}} & 515,345 & 463,715 & 51,630 & 91  \\ \hline
  \end{tabular}}
\end{table}

We divide the observations in  \textit{Airfoil}, \textit{Parkinson}, and \textit{Protein} into training and test sets by extracting $85\%$ of the sample as training and the rest as test points. In the other data sets, there are predefined training and test sets. Indeed, in \textit{Song} data set, we extract the first $10^5$ songs from this data set for training and the first $10^4$ songs from the original test set for testing. We used disjoint partitioning (K-means) to divide the data sets into 5(for \textit{Airfoil}), 10(for \textit{Parkinson} and \textit{Pole Telecom}), 70 (for \textit{Protein}), 72 (for \textit{Sacros}), and 80 (for \textit{Song}) subsets.

Next, we compare SOTA baselines with classification-based aggregations. For the selection-based methods, we set $K_M$ to $0.5$, which means $50\%$ of experts are selected. For the \textit{Song} data set only, we set $K_M=0.2$ in. Since NPAE is computationally burdensome, especially when $M$ and $N_t$ are large, it is only used in small and medium-scale data sets. DNN uses a neural network with a hidden layer and 50 hidden units to evaluate the labels. 

\begin{table*}[hbt!]
\caption{\textbf{SMSE} for various methods on real-world data sets. The table depicts SMSE values for SOTA  baselines and the classification-based aggregations, i.e. KNN, and DNN. Both dependent (D) and conditionally independent (CI) aggregation methods are used.}
\label{table.SMSE_real_data_experiments}
\vskip -0.3in
\begin{center}
\begin{small}
\begin{sc}
\scalebox{1.2}{\begin{tabular}{lcccccr}
\toprule
\multirow{1}{*}{} &
\multicolumn{2}{c}{} &
\multicolumn{2}{c}{SMSE} &
\multicolumn{2}{c}{} \\
\toprule
 Model &\textit{Airfoil} & \textit{Parkinson}  & \textit{Pole Telecom}  & \textit{Protein} & \textit{Sacros} & \textit{Song}  \\
\midrule
GPoE (CI) &0.1305& 0.2703 & 0.0727& 0.8654& 0.0461& 0.9221  \\
RBCM (CI) &0.0881& 0.2339& 0.0237 & 0.3569 & 0.0039& 0.8127 \\
GRBCM (CI) &0.0777& 0.2395& 0.0191 & 0.3540 & 0.0034& 0.7762  \\
\midrule
NPAE (D) & \textbf{0.0694} & \textbf{0.2121} & \textbf{0.0144} &\textbf{-}& \textbf{-} &\textbf{-}\\
\midrule
\textcolor{blue}{GMM-5} (D) & 0.0765 & 0.2317 & 0.0182& 0.3326& \textbf{0.0025} & 0.7379 \\
\textcolor{blue}{KNN-5} (D) & \textbf{0.0694} & \textbf{0.2126} & \textbf{0.0145} & \textbf{0.2743} & \textbf{0.0025} & \textbf{0.7007} \\
\textcolor{blue}{DNN-5} (D) & \textbf{0.0694} & \textbf{0.2127} & \textbf{0.0141} & \textbf{0.2744} & \textbf{0.0025} & \textbf{0.6994} \\
\bottomrule
\end{tabular}}
\end{sc}
\end{small}
\end{center}
\vskip 0.1in
\end{table*}

\begin{table*}[hbt!]
\caption{\textbf{MSLL} for various methods on real-world data sets. The table depicts MSLL values for SOTA  baselines and the classification-based aggregations, i.e. KNN, and DNN. Both dependent (D) and conditionally independent (CI) aggregation methods are used.}
\label{table.MSLL_real_data_experiments}
\vskip -0.3in
\begin{center}
\begin{small}
\begin{sc}
\scalebox{1.2}{\begin{tabular}{lcccccr}
\toprule
\multirow{1}{*}{} &
\multicolumn{2}{c}{} &
\multicolumn{2}{c}{MSLL} &
\multicolumn{2}{c}{} \\
\toprule
 Model &\textit{Airfoil} & \textit{Parkinson}  & \textit{Pole Telecom}  & \textit{Protein} & \textit{Sacros} & \textit{Song}  \\
\midrule
GPoE (CI) &-1.1875& -0.5862 & -1.5171& -0.0759& -1.165& -0.0449  \\
RBCM (CI) &-1.3187& -0.5433 & -1.5901 & -0.6164 & -2.5347 & -0.1266 \\
GRBCM (CI) & -1.4706 & -0.8123 & -2.293 & -0.6378 & -2.7985& -0.1563  \\
\midrule
NPAE (D) & \textbf{-1.5207} & \textbf{-0.8583} & -2.3537 &\textbf{-}& \textbf{-} &\textbf{-}\\
\midrule
\textcolor{blue}{GMM-5} (D) & -1.4928 & -0.7937 & -2.1658 & -0.6173 & \textbf{-2.8017} & -0.1613 \\
\textcolor{blue}{KNN-5} (D) & \textbf{-1.5209} & \textbf{-0.8563} & \textbf{-2.3851} & \textbf{-0.7348} & -2.8005 & \textbf{-0.1913} \\
\textcolor{blue}{DNN-5} (D) & \textbf{-1.5208} & \textbf{-0.8569} & \textbf{-2.3823} & \textbf{-0.7349} & \textbf{-2.8023} & \textbf{-0.1926} \\
\bottomrule
\end{tabular}}
\end{sc}
\end{small}
\end{center}
\vskip -0.1in
\end{table*}

Tables \ref{table.SMSE_real_data_experiments} and \ref{table.MSLL_real_data_experiments} reveal the SMSE and MSLL values of the baselines. Selecting the experts enables us to encode dependency between agents efficiently while the prediction quality is comparable with the original baseline NPAE. However, their running times are acceptable, especially when dealing with high-dimensional large data sets where using the NPAE is not feasible. Besides, the convergence rate of KNN and DNN is much faster than GGM. Even with 50\% of experts, the results of the GGM still have a remarkable deviation from NPAE and cannot provide relevant results in the different data sets. 

We consider \textit{Parkinson} data set as an instance. GGM can not provide an accurate prediction when $K_M=0.5$. Our experiments with $K_M=0.7$ \footnote{The results for $K_M=0.7$ are not shown in the Tables \ref{table.SMSE_real_data_experiments} and \ref{table.MSLL_real_data_experiments}} confirm that increasing the $K_M$ to $0.7$ for GGM reduces the SMSE to $0.2208$ and MSLL to $-0.8541$ which are close to the SMSE and MSLL of NPAE. It indicates that GGM converges to NPAE, but the convergence is slower than KNN and DNN. Meanwhile, the SMSE values of the KNN and DNN methods for $K_M=0.7$ are $0.2122$ and $0.2117$, respectively\footnote{the MSLL values of the KNN and DNN methods for $K_M=0.7$ are $-0.8575$ and $-0.8587$, respectively.}. Therefore, by including more experts in the final aggregation, KNN, and DNN can outperform the NPAE by excluding the effects of low-quality experts in Equation \eqref{eq:npae}.

In CI-based methods, the conservative GPoE does not return acceptable results and can not outperform RBCM and GRBCM methods. The quality of the GRBCM is slightly better than RBCM because of the global communication expert. The global expert improves the quality measures of GRBCM compared to RBCM, especially in MSLL, where the values are always smaller the those of RBCM. Both methods provide competitive results with GGM when $K_M=0.5$. However, their deviation from NPAE, KNN, and DNN is remarkable. Indeed, by increasing the $K_M$ GGM can easily outperform them.  
\section{Conclusion} \label{conclusion}
In this work, we have proposed a novel expert selection approach for distributed learning with Gaussian agents, which leverages expert selection to aggregate dependent local experts' predictions. The available ensemble baselines use all correlated experts in the aggregation step. It affects the final predictions by the local predictions of weak experts or leads to impractically high computational costs. Our proposed approach uses a multi-label classification model, and the allocation of data points to experts is defined by considering the experts as class labels. 

Unlike the available expert selection method that assigns a fixed and static set of the selected experts to all new data points, the proposed model is more flexible to the model and data changes and chooses a related group of experts at each entry point. Excluding unrelated experts at each test point improves the prediction quality and reduces computational costs. Meanwhile, it keeps the original baseline's asymptotic properties that use all exerts and provides consistent results when $n \to \infty$. The classification methods in this work, i.e., KNN and DNN, can be replaced with recent and more efficient solutions proposed to solve the multi-label classification problem. The proposed approach can be used for distributed and federated learning and does not impose restricted assumptions. Through empirical analyses, we illustrated the superiority of our approach, which improves the prediction quality of existing SOTA aggregation methods while being highly efficient.

\bibliographystyle{IEEEtran}
\bibliography{IEEEabrv,ref}

\vfill

\end{document}